\documentclass{article} 
\usepackage[preprint]{colm2026_conference}

\usepackage{microtype}
\usepackage{hyperref}
\usepackage{url}
\usepackage{booktabs}

\usepackage{lineno}

\usepackage[utf8]{inputenc} 
\usepackage[T1]{fontenc}    
\usepackage{hyperref}       
\usepackage{url}            
\usepackage{booktabs}       
\usepackage{amsfonts}       
\usepackage{nicefrac}       
\usepackage{microtype}      
\usepackage{xcolor}         
\usepackage[table]{xcolor}         
\usepackage{algorithm}
\usepackage{algorithmic}
\usepackage{lipsum}
\usepackage{caption}

\usepackage{enumitem}

\usepackage{amssymb}
\usepackage{amsmath}

\usepackage{amsthm}
\usepackage{url}
\usepackage{graphicx}
\usepackage{multirow}
\usepackage{multicol}
\usepackage{makecell}
\usepackage{subcaption}
\usepackage{wrapfig}

\usepackage{fancyvrb}
\usepackage{listings}
\usepackage{longtable}   
\usepackage{array} 
\usepackage{wrapfig} 

\theoremstyle{plain}
\newtheorem{definition}{Definition}

\newcommand{\latinabbrev}[1]{\textit{#1}}
\newcommand{\eg}{\latinabbrev{e.g.}}

\newcommand{\etc}{\latinabbrev{etc}}
\newcommand{\ie}{\latinabbrev{i.e.}}

\newcommand{\fname}{\textsc{MetaSymbO}}

\newcommand{\algcomment}[1]{\hfill$\triangleright$~#1}

\definecolor{darkblue}{rgb}{0, 0, 0.5}
\hypersetup{colorlinks=true, citecolor=darkblue, linkcolor=darkblue, urlcolor=darkblue}

\title{\fname: Multi-Agent Language-Guided Metamaterial Discovery via Symbolic Latent Evolution}

\author{
Jianpeng Chen\textsuperscript{1},
Wangzhi Zhan\textsuperscript{1},
Dongqi Fu\textsuperscript{2},
Junkai Zhang\textsuperscript{3},
Zian Jia\textsuperscript{4,5},
Ling Li\textsuperscript{5}
\\
\textbf{
Wei Wang\textsuperscript{3},
Dawei Zhou\textsuperscript{1}
}\\
\textsuperscript{1}Virginia Tech \quad
\textsuperscript{2}Meta AI \quad
\textsuperscript{3}University of California Los Angeles \\
\textsuperscript{4}Princeton University \quad
\textsuperscript{5}University of Pennsylvania\\
jianpengc@vt.edu
}

\begin{document}

\ifcolmsubmission
\linenumbers
\fi

\maketitle

\begin{abstract}
Metamaterial discovery seeks microstructured materials whose geometry induces targeted mechanical behavior. Existing inverse-design methods can efficiently generate candidates, but they typically require explicit numerical property targets and are less suitable for early-stage exploration, where researchers often begin with incomplete constraints and qualitative intents expressed in natural language. Large language models can interpret such intents, but they lack geometric awareness and physical property validity. To address this gap, we propose \textbf{\fname}, a multi-agent framework for language-guided \textbf{Meta}material discovery via \textbf{Symb}olic-driven latent Ev\textbf{O}lution. Specifically, \fname\ contains three agents: a Designer that interprets free-form design intents and retrieves a semantically consistent \textit{scaffold}, a Generator that synthesizes candidate microstructures in a disentangled latent space, and a Supervisor that provides fast property-aware feedback for iterative refinement. To move beyond the limitations of reproducing known samples from literature and training data, we further introduce symbolic-driven latent evolution, which applies programmable operators over disentangled latent factors to compose, modify, and refine structures at inference time.
Extensive experiments demonstrate that (i) \fname\ improves structural validity by up to 34\% in symmetry and nearly 98\% in periodicity compared to state-of-the-art baselines; (ii) \fname\ achieves about 6–7\% higher language-guidance scores while maintaining superior structure novelty compared to advanced reasoning LLMs; (iii) qualitative analyses confirm the effectiveness of symbolic logic operators in enabling programmable semantic alignment; and (iv) real-world case studies on auxetic, high-stiffness metamaterial design further validate its practical capability. 
\end{abstract}
\vspace{-1em}
\section{Introduction}
\vspace{-1em}
Metamaterials, an emerging microstructured category of materials, are receiving increasing attention due to their capability to achieve extraordinary mechanical properties, 
exhibiting wide applications in various fields, such as biomedical devices, transportation systems, robotics, \etc.~\citep{paul2010optical,engheta2006metamaterials,jia2020engineering}.
A central goal of metamaterial discovery is to identify microstructures with desired mechanical responses, such as target elastic moduli or Poisson’s ratios~\citep{RW006}. 
As the design space of possible microstructures is extremely large, recent research has increasingly shifted toward data-driven methods to improve the efficiency of metamaterial discovery. In particular, variational autoencoders (VAEs)~\citep{vae}, diffusion models (DMs)~\citep{podell2023sdxl,fu2024latent,zhan2025unimate}, and generative adversarial networks (GANs)~\citep{RW001,RW003} have been applied to learn structure--property relationships for inverse metamaterial design tasks~\citep{uniTruss,CDVAE,RW005}. However, although these methods are effective for exploring large design spaces, they usually require explicit numerical property specifications and cannot naturally interpret free-form language descriptions or implicit domain knowledge from the literature, which are often critical in the early stage of metamaterial design.

Recent studies~\citep{RW008,chen2024generative} indicate that metamaterial discovery often begins with incomplete information, evolving constraints, and only vague conceptual goals. Traditional methods which rely on explicit numerical property inputs struggle in such exploratory phases and can even fail to produce valid outputs~\citep{jin2023mechanical,ronellenfitsch2019inverse}. By contrast, natural language provides a flexible way to specify qualitative design intents (\eg, ``lightweight and energy-absorbing under impact'') without committing to precise targets, making it well-suited for exploratory workflows. Large language models (LLMs) further extend this flexibility by their strong capabilities in understanding language and retrieving domain knowledge~\citep{tian2025multi,narayanan2025aviary}, as demonstrated by LLMs such as GPT-4o~\citep{gpt-4o} and DeepSeek~\citep{guo2025deepseek}. Nevertheless, this advantage remains limited by the fact that LLMs are not inherently geometry-aware and cannot explicitly enforce physical constraints. Consequently, they tend to either retrieve existing designs from the literature or produce geometries that are physically invalid.

These observations reveal a persistent modality gap: LLM agents are strong at interpreting language-level design intents, whereas geometry-aware generators are effective at producing structurally valid and physically realistic designs, but neither alone can bridge these domains. Therefore, a natural research question arises, 
\textit{can we build a unified metamaterial scientist, which has \textbf{multiple domain expertise} in \textbf{geometric topology awareness}, \textbf{flexible natural language understanding}, and \textbf{effective design capability}, for metamaterial discovery}? 

To achieve this, we identify two critical challenges.
\textbf{C1:} \textit{How can we bridge the modality gap, considering the large discrepancy among modalities?} Language-guided metamaterial discovery involves three distinct modalities: language, geometry, and properties. For language modality, interpreting qualitative design intents, such as ``strong but flexible'', requires robust reasoning capabilities and substantial domain expertise. For geometric modality, designing microstructures demands an understanding of structural consistency and adherence to physical constraints. For property modality, creating materials to achieve targeted properties requires comprehensive knowledge of the mechanical properties.
\textbf{C2:} \textit{How can we expand the existing design space to enable broader hypothesis exploration?} Both LLMs and generative models typically operate within existing literatures or training data design spaces. However, the metamaterial design targets often exceed the design space of them. Exploring unknown hypotheses beyond these initial spaces remains an important challenge.

To address the two challenges, we propose \textbf{\fname}, a multi-agent collaborative framework utilizing symbolic-driven latent evolution for metamaterial discovery. Specifically, for \textbf{C1}, we introduce three specialized agents, each focusing on one modality. Agent Designer, powered by an LLM, provides language reasoning to interpret prompts and query the domain literature. Agent Generator is a generative model specialized in geometric modality, enabling exploration within vast valid geometric design spaces. Agent Supervisor integrates property prediction and language reasoning, facilitating property-driven design. Moreover, a collaboration mechanism is developed for joint structure optimization. For \textbf{C2}, motivated by the programmable methods~\citep{bastek2022inverting,ZHAO2026106351}, we propose a symbolic-driven latent evolution approach to optimize geometries toward language semantics in Generator, achieving effective hypothesis exploration outside the literature domain and the training data domain.
The contributions can be summarized:
\begin{itemize}[leftmargin=*,topsep=0pt]
    \item \textbf{Conceptual Contribution:}
    We unify language guidance, geometric generation, and property-aware reasoning in a single pipeline to enable the practical metamaterial discovery workflows starting with a conceptual-level idea.
    \item \textbf{Methodology Contribution:}
    We propose \fname, a multi-agent framework for efficient and robust metamaterial discovery, together with a symbolic-driven latent evolution module for inference-time cross-domain exploration.
    \item \textbf{Benchmark Contribution:}
    We benchmark \fname\ against state-of-the-art LLMs and generative models, showing superior structural validity, design diversity, and language-guidance effectiveness. These quantitative comparisons validate the effectiveness.
    \item \textbf{Practical Contribution:}
    Case studies demonstrate the end-to-end applicability of \fname\ in realistic metamaterial design workflows, from prompt refinement and structure generation to downstream physical validation of 3D printed samples.
\end{itemize}

\section{Preliminary}
\vspace{-0.5em}
\paragraph{Metamaterial Design.}
\begin{wrapfigure}{r}{10em}
\small
    \vspace{-3em}
    \includegraphics[width=\linewidth]{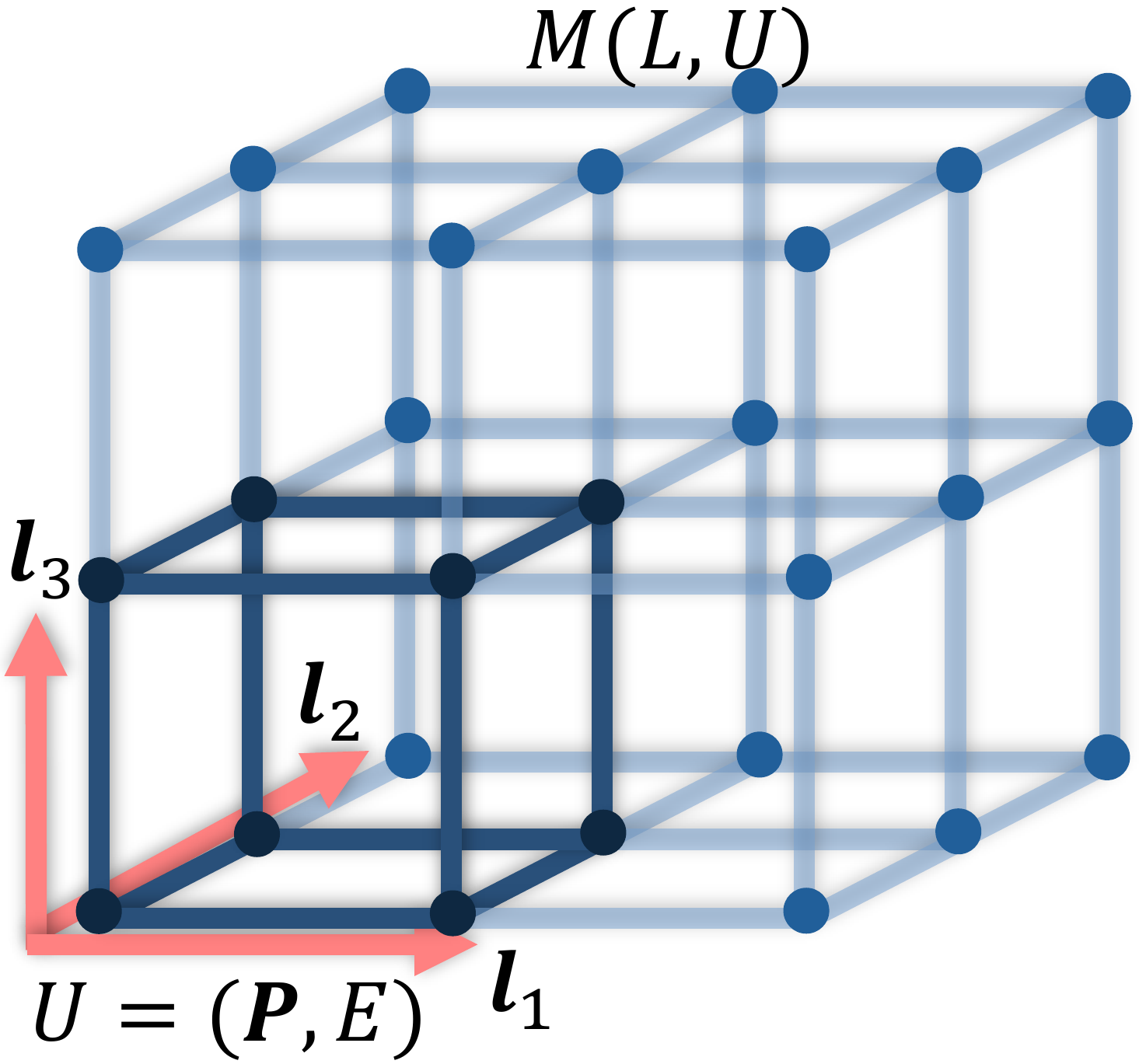}
    \vspace{-1.5em}
    \caption{Truss lattice.}
    \vspace{-2em}
    \label{fig:Lattice}
\end{wrapfigure}
Metamaterials, as shown in Figure~\ref{fig:Lattice}, are typically modeled as lattice structures, \ie, periodic arrangements of unit cells composed of trusses in geometric space~\citep{chen2025metamatbench}.
Formally, a metamaterial can be represented as $M=(\mathbf{L},\mathcal{U})$, each unit cell $\mathcal{U}=(\mathbf{P},E)$ consists of node coordinates $\mathbf{P}\in\mathbb{R}^{N\times3}$ and edge set $E$ specifying strut connections, and the lattice vectors $\mathbf{L}=[\boldsymbol{l}_1, \boldsymbol{l}_2, \boldsymbol{l}_3]^\mathrm{T}\in\mathbb{R}^{3\times3}$ define the periodic tiling of unit cells in 3D space.
The corresponding mechanical properties are denoted by $\mathbf{y}\in\mathbb{R}^{d_y}$.
The central goal of metamaterial discovery is to design structures with tailored properties that satisfy specific functional requirements.
\vspace{-0.5em}
\paragraph{Limitations of Existing Works.}
Recent metamaterial discovery has increasingly relied on data-driven methods that learn structure--property relationships from data. Generative models such as VAEs and DMs typically optimize a conditional autoencoder $q_{\boldsymbol{\phi}}(\mathbf{z}\mid M)$ and $p_{\boldsymbol{\theta}}(M\mid \mathbf{z},\mathbf{y})$ under a Gaussian prior $p(\mathbf{z})\sim\mathcal{N}(\mathbf{0},\mathbf{I})$. While effective for inverse design, these methods usually depend on explicit numerical property conditions (\ie, $\mathbf{y}$) and often entangle geometric and semantic information in a shared latent space, limiting controllability and generation quality.

LLM-based approaches are trained on large volumes of domain literature and therefore offer strong potential for retrieving candidates from broad design spaces~\citep{esm3,CrystaLLM,EquiLLM,Geo2Seq,bindgpt,narayanan2025aviary}. However, existing methods either adapt LLMs to strictly formatted material representations, which weakens their ability to understand free-form language, or simply prompt them to generate materials directly, causing them to favor reproducing known materials from the literature rather than proposing novel designs.

\vspace{-0.5em}
\paragraph{Problem Statement.} Given a free-form language prompt, the goal of this work is to perform language-guided metamaterial design to generate structures whose mechanical properties satisfy the intent expressed in the prompt.
\vspace{-0.7em}
\section{\fname: Language-Guided Metamaterial Discovery}
\begin{figure*}
\small
    \centering
    \vspace{-2em}
    \includegraphics[width=\linewidth]{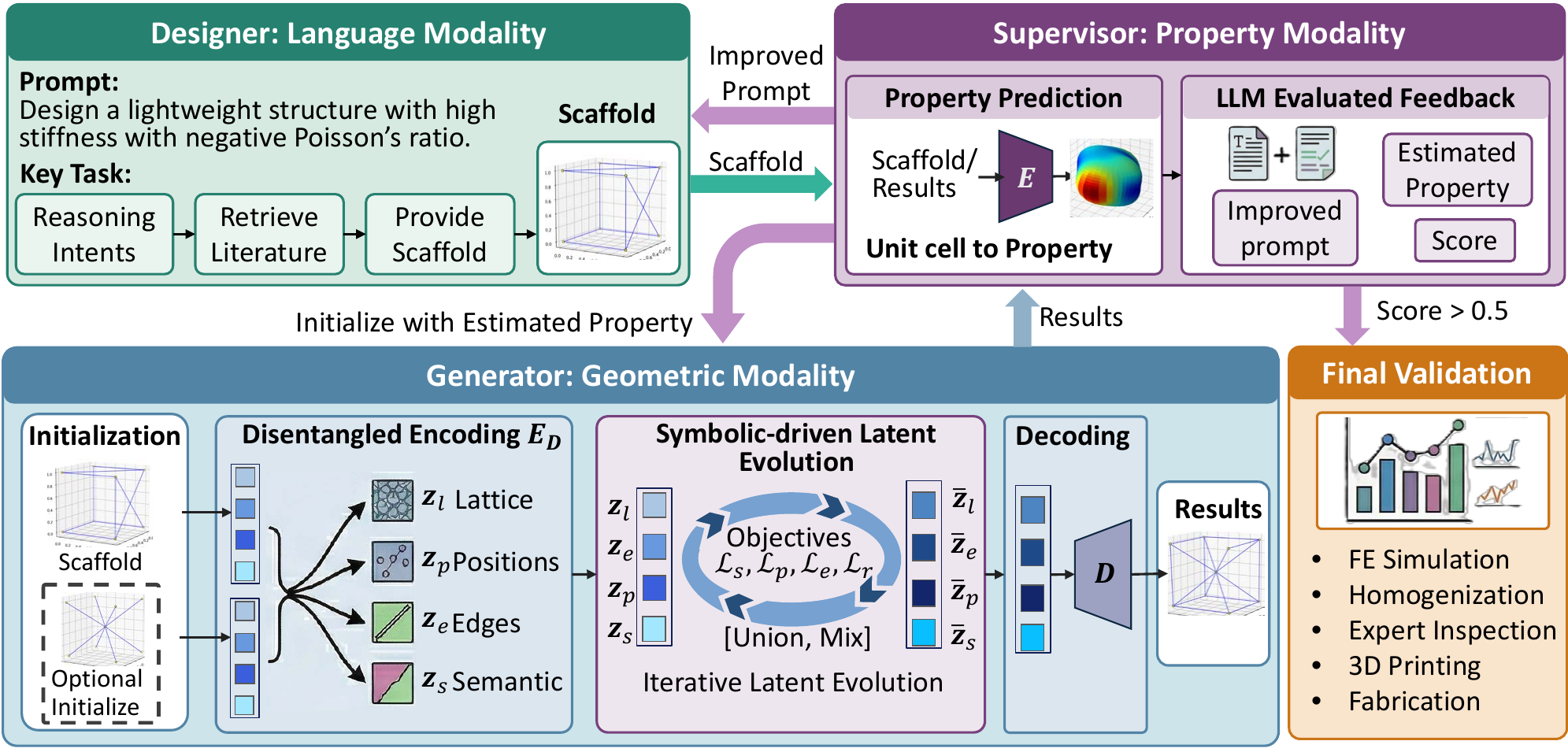}
    \vspace{-1.5em}
    \caption{\textbf{Overview of \fname}. Agent Designer translates the prompt into a scaffold, Agent Generator refines the design in latent geometric space via symbolic-driven latent evolution, and Agent Supervisor evaluates properties to provide fast iterative feedback.}
    \label{fig:framework}
    \vspace{-1em}
\end{figure*}

\vspace{-0.7em}
\label{sec:agents_intro}
In this section, we present \fname, a multi-agent framework for language-guided metamaterial discovery, as shown in Figure~\ref{fig:framework}. To bridge the modality gap across language, geometry, and properties (Challenge \textbf{C1}), \fname\ coordinates three specialized agents. \textbf{Designer} interprets the input prompt and retrieves a simple existing structure (called as \textit{scaffold}), which serves as a semantic anchor between free-form language and geometric generation. Guided by this scaffold, \textbf{Generator} searches the design space in the proposed disentangled latent space, where semantic and geometric factors are explicitly separated for controllable and valid synthesis. To further enable exploration beyond known designs (Challenge \textbf{C2}), we introduce \textit{symbolic-driven latent evolution}, which refines latent representations to align with scaffold semantics while expanding the reachable design space. Finally, \textbf{Supervisor} provides fast property prediction and language-based evaluation to assess generated candidates for iterative feedback, forming a closed-loop optimization process for language-guided metamaterial design.

\vspace{-0.5em}
\subsection{Designer: Language‑Modality Designer}
\label{sec:agent_1}
\vspace{-0.5em}
To address the modality gap between conceptual-level language and complex geometric space, there are two central issues: (1) the initial metamaterial design idea might be vague and incomplete~\citep{RW008,chen2024generative}, which requires the Designer to be equipped with a large amount of domain literature to infer the intent from the initial prompt; (2) after the Designer could successfully understand the prompt, the critical issue is how the model can translate the intent of prompt into geometric space.

Fortunately, LLMs have been demonstrated to be effective for understanding, retrieving, and reasoning in language space for domain-specific scientific discovery. These previous works demonstrate the capability of LLMs to infer the intent from a prompt: for example, the phrase ``a hard material'' statistically co‑occurs with ``high Young’s modulus'' in the literature, allowing the model to map the qualitative adjective ``hard’’ to a quantitative stiffness concept~\citep{jin2023mechanical} (as examples validated in Appendix~\ref{app:vali_agent1}).
As for the second issue, although LLMs can render existing classic lattices, they often fail when it comes to more complex generation tasks and larger geometric design spaces. The experiments for LLMs in Table~\ref{tab:quant_comp} show that LLMs tend to generate repeated structures, indicating their capability in successfully retrieving existing structures and limitations to explore large geometric design space. To balance this, we propose ``scaffold''.

\textbf{Scaffold}, the output of Designer, refers to existing simple geometric structures that imply the core intent of the initial prompt. The high-level idea of it is to utilize the strong retrieval and language-understanding power of LLMs. For example, given a prompt for generating ``stable'' structure, we expect LLMs to retrieve a scaffold of an octet or triangle-like structure (as shown in Appendix~\ref{app:vali_agent1}), rather than a cubic that is less stable. The model thus outputs a concise scaffold description that is semantically consistent with the prompt.
\vspace{-0.5em}
\subsection{Generator: Geometry‑Modality Synthesizer}\label{sec:agent_2}
\vspace{-0.5em}
Given a scaffold that reflects the semantics of the initial design concept, Generator operates in geometric modality to solve two problems: semantic grounding and structural extrapolation. Specifically, it should align generated geometries with scaffold semantics to bridge the language–geometry modality gap (\textbf{C1}), while also exploring candidate hypotheses beyond the support of the training distribution (\textbf{C2}).

We therefore introduce a \textbf{symbolic-driven latent evolution} method in a disentangled latent space to address both challenges. To tackle \textbf{C1}, the proposed method injects scaffold semantics into geometric generation, bridging high-level language intent and geometric generation. To tackle \textbf{C2}, motivated by programmable metamaterial design strategies that derive novel structures through the composition and transformation of existing motifs~\citep{bastek2022inverting,ZHAO2026106351}, our method evolves and recombines scaffold representations to construct new designs beyond existing examples. Moreover, performing symbolic operations in a disentangled latent space improves generation quality while enabling semantic-level integration.

Accordingly, Agent Generator follows four stages, as shown in Figure~\ref{fig:framework}: initialization from scaffold and optional conditions, disentangled latent encoding, symbolic-driven latent evolution, and decoding. This design enables inference-time semantic programming of geometry while preserving the flexibility of data-driven generation. We describe more details below.
\vspace{-0.5em}
\paragraph{Latent Disentanglement.}
Existing latent generative frameworks~\citep{CDVAE,condCDVAE,SyMat} typically use a unified latent space, where different metamaterial attributes, \ie, node positions $\mathbf{P}$, edge connectivity $E$, lattice vectors $\mathbf{L}$, and properties $\mathbf{y}$, are encoded in a single Gaussian representation. Such entanglement limits controllability and generation quality. To enable finer-grained control and semantic alignment, we propose to disentangle the latent variable. 
Given a metamaterial $\mathcal{M}=(\mathbf{L},\mathcal{U})$ with $\mathcal{U}=(\mathbf{P},E)$ and associated properties $\mathbf{y}$, we decompose the latent variable $\mathbf{z}$ into four components,
$\mathbf{z}_l, \mathbf{z}_p, \mathbf{z}_e, \mathbf{z}_s$,
corresponding to lattice, node positions, edges, and semantic/property factors, respectively. The inference and generation processes are defined as:
\begin{equation}\label{eq_main:enc}
\small{
\begin{aligned}
&p_\theta(\mathcal{M}, \mathbf{y}, \mathbf{z})
= p(\mathbf{z})\,
p_{\theta_l}(\mathbf{L}\mid \mathbf{z}_l,\mathbf{y})\,
p_{\theta_p}(\mathbf{P}\mid \mathbf{z}_p,\mathbf{y})\,
p_{\theta_e}(E\mid \mathbf{z}_e,\mathbf{y})\,
p_{\theta_s}(\mathbf{y}\mid \mathbf{z}_s), \\
&q_\phi(\mathbf{z}\mid \mathcal{M})
= q_\phi(\mathbf{z}_l,\mathbf{z}_p,\mathbf{z}_e,\mathbf{z}_s \mid \mathcal{M}) \\
&\qquad = q_{\phi_1}(\mathbf{z}_l\mid \mathcal{M})\,
q_{\phi_2}(\mathbf{z}_p\mid \mathcal{M})\,
q_{\phi_3}(\mathbf{z}_e\mid \mathcal{M})\,
q_{\phi_4}(\mathbf{z}_s\mid \mathcal{M}), \\
&p(\mathbf{z}) = p(\mathbf{z}_l)\,p(\mathbf{z}_p)\,p(\mathbf{z}_e)\,p(\mathbf{z}_s),
\qquad p(\cdot)\sim\mathcal{N}(\mathbf{0},\mathbf{I}).
\end{aligned}
}
\end{equation}
This disentanglement allows Generator to manipulate structural and semantic factors separately, which is critical for scaffold-guided synthesis.
\vspace{-0.5em}
\paragraph{Symbolic-Driven Latent Evolution.}
To preserve and program the semantics implied by the prompt, Generator should align the latent representation of an initialized structure with that of the scaffold. We focus on two symbolic operators in the main text---\textit{Union} and \textit{Mix}---as they capture two intuitive forms of scaffold-guided programming: structural expansion and semantic interpolation. Additional operators, including \textit{Intersection} (Eq.~\eqref{eq:intersection}) and \textit{Negation} (Eq.~\eqref{eq:neg}), are deferred to Appendix~\ref{app:symbo_opt}.

We first initialize a metamaterial $\mathcal{M}$ from training dataset. \textit{Union} aims to unite the scaffold (represented as $\mathcal{M}'$) with initialized sample. 
Concretely, we compute a soft node correspondence between $\mathcal{M}$ and $\mathcal{M}'$ using Sinkhorn matching~\citep{frogner2015learning} (Eq.~\eqref{eq:union}), which identifies overlapping components and enables semantic fusion at the node level. Union is therefore suitable when the scaffold provides missing structural elements that should be incorporated into the current design.

In contrast, \textit{Mix} performs continuous semantic interpolation between the initialized structure and the scaffold in latent space. Let $p_{\mathcal{M}}(\mathbf{z})$ and $p_{\mathcal{M}'}(\mathbf{z})$ denote their latent distributions. We define
\begin{equation}
p_{\mathrm{mix}}(\mathbf{z}\mid \lambda_{\mathrm{mix}})
=
(1-\lambda_{\mathrm{mix}})\,p_{\mathcal{M}}(\mathbf{z})
+
\lambda_{\mathrm{mix}}\,p_{\mathcal{M}'}(\mathbf{z}),
\qquad
\lambda_{\mathrm{mix}}\in[0,1],
\end{equation}
where $\lambda_{\mathrm{mix}}$ controls the degree to which the scaffold influences the resulting design. Smaller values preserve more of the initialization, while larger values pull the latent representation closer to the scaffold semantics. The final expression is shown in Appendix~Eq.~\eqref{eq:mixture}.

With the target latent distributions derived from the symbolic operations, we then propsoe to evolve the generation latent towards the target latent distribution for two reasons. First, in a disentangled autoencoder, the decoder is trained only on the latent manifold induced by the encoder; operator-based targets may fall outside this manifold and thus lead to invalid decoded structures. Second, symbolic fusion is applied within individual subspaces, disturbing the compatibility among $\mathbf{z}_l$, $\mathbf{z}_p$, $\mathbf{z}_e$, and $\mathbf{z}_s$ that the decoder implicitly relies on.

To do so, we evolve the latents of the initialized structure toward the operator-induced target via gradient descent, rather than decoding the target distribution directly. During optimization, we impose a Sinkhorn-based soft-matching constraint on node and edge distributions to preserve cross-space coherence and keep the latent trajectory close to the learned manifold. Let $\boldsymbol{\mu}$ and $\boldsymbol{\sigma}$ denote Gaussian means and variances. The optimization objective is
\begin{equation}\label{eq_main:objective}
\small{
\begin{aligned}
&\mathcal{L}_{s}
= \mathrm{KL}\!\left(
\mathcal{N}(\boldsymbol{\mu}_{s},\boldsymbol{\sigma}_{s})
\;\|\;
\mathcal{N}(\boldsymbol{\mu}'_{s},\boldsymbol{\sigma}'_{s})
\right)
\qquad \text{(semantic alignment)},\\
&\mathcal{L}_{p,e}
= \sum_{k\in\{p,e\}}
\sum_{i=1}^{N_{\mathcal{M}}}
\sum_{j=1}^{N_{\mathcal{M}'}}
\Pi_{ij}\,
\mathrm{KL}\!\left(
\mathcal{N}(\boldsymbol{\mu}_{k,i},\boldsymbol{\sigma}_{k,i})
\;\|\;
\mathcal{N}(\boldsymbol{\mu}'_{k,j},\boldsymbol{\sigma}'_{k,j})
\right)
\qquad \text{(Sinkhorn-weighted alignment)},\\
&\mathcal{L}_{r}
= \sum_{k\in\{p,e\}}
\sum_{i\in\{i\mid r_i<\tau_o\}}
\mathrm{KL}\!\left(
\mathcal{N}(\boldsymbol{\mu}_{k,i},\boldsymbol{\sigma}_{k,i})
\;\|\;
\mathcal{N}(\boldsymbol{\mu}^{\mathrm{old}}_{k,i},\boldsymbol{\sigma}^{\mathrm{old}}_{k,i})
\right)
\qquad \text{(unmatched-node regularization)},\\
&\mathcal{L}_{\mathrm{prior}}
= \sum_{i\in\{l,p,e,s\}}\|\mathbf{z}_i\|_2^2
\qquad \text{(latent prior regularization)}.
\end{aligned}
}
\end{equation}
Here, $\Pi_{ij}$ denotes the Sinkhorn soft-matching matrix between nodes of $\mathcal{M}$ and $\mathcal{M}'$, $\mathcal{N}(\boldsymbol{\mu}^{\mathrm{old}}_{k,i},\boldsymbol{\sigma}^{\mathrm{old}}_{k,i})$ denotes the pre-optimization distribution, and $\mathcal{N}(\boldsymbol{\mu}',\boldsymbol{\sigma}')$ denotes the operator-induced target. The set $\{i\mid r_i<\tau_o\}$ contains unmatched nodes under Sinkhorn matching (Appendix~Alg.~\ref{alg:sinkhorn_log}), with threshold $\tau_o=0.1$. More implementation details and mathematical derivations are provided in Appendix~\ref{app:symbo_opt}.

\vspace{-0.7em}
\subsection{Supervisor: Property‑Modality Evaluator}\label{sec:agent_3}
\vspace{-0.7em}
\begin{algorithm}[b]
\caption{\footnotesize\textbf{Designer\&Supervisor and Generator\&Supervisor Collaborations} \\
$\text{Init}(\mathbf{z} \mid \mathbf{y})$ represents initializing $\mathbf{z}$ given property $\mathbf{y}$, which is instantiated as a $1$-nearest-neighbor~\citep{knn} from the dataset followed by Gaussian latent encoding. $D,G,S$ denote Designer, Generator, and Supervisor. $S^{\text{pred}}$ and $S^{\text{eval}}$ are the predictor and LLM in Supervisor.}
\label{alg:collaborations}
\footnotesize
\noindent
\begin{minipage}[t]{0.48\linewidth}
\begin{algorithmic}
\STATE \textbf{$\boldsymbol{C_{D/S}}$ (Scaffold refinement)}
\\\textbf{Input:} prompt $V_p$; \textbf{Output:} scaffold $V_m$.
\STATE Step1: $V_m^{(t)} \gets D(V_p^{(t)})$
\STATE Step 2: $\mathbf{y}_s^{(t)} \gets S^{\text{pred}}\!\Big(f_{\mathcal{V}/\mathcal{M}}(V_m^{(t)})\Big)$
\STATE Step 3: $score^{(t)}, V_p^{(t+1)} \gets S^{\text{eval}}\!\Big(V_m^{(t)}, V_p^{(t)}, \mathbf{y}_s^{(t)}\Big)$
\STATE Step 4: \textbf{repeat} until $score^{(t)} \ge \tau_{D/S}$
\end{algorithmic}
\end{minipage}
\hfill
\begin{minipage}[t]{0.6\linewidth}
\begin{algorithmic}
\STATE \textbf{$\boldsymbol{C_{G/S}}$ (Generation refinement)}
\\\textbf{Input:} scaffold $V_m$; \textbf{Output:} microstructure $\overline{M}$.
\STATE Step 1: $\overline{M}^{(t)} \gets G(\mathbf{z}^{(t)},f_{V/M}(V_m))$, $\mathbf{z}^{(0)} \sim \mathcal{N}(\mathbf{0}, \mathbf{I})$
\STATE Step 2: $\mathbf{y}_m^{(t)} \gets S^{\text{pred}}(\overline{M}^{(t)})$
\STATE Step 3: $score^{(t)}, \mathbf{y}_m^{\prime(t)} \gets S^{\text{eval}}\!\Big(f_{\mathcal{M}/\mathcal{V}}(\overline{M}^{(t)}), V_p, \mathbf{y}_m{(t)}\Big)$
\STATE Step 4: $\mathbf{z}^{(t+1)} \gets \textbf{Init}\!\big(\mathbf{z}^{(t)} \mid \mathbf{y}_m^{\prime(t)}\big)$ {optional reinitialize}
\STATE Step 5: \textbf{repeat} until $score^{(t)} \ge \tau_{G/S}$
\end{algorithmic}
\end{minipage}
\end{algorithm}
Designer and Generator provide language-space reasoning and geometry-space synthesis, respectively, but automated metamaterial discovery still requires efficient feedback in property space. Directly conducting high-fidelity simulation inside the design loop is computationally prohibitive~\citep{lee2024data}. Therefore, \textbf{Supervisor} is a fast property-space evaluator that provides approximate yet informative feedback for iterative refinement.

Specifically, Supervisor consists of two components as shown in Figure~\ref{fig:framework}: (1) a property predictor trained on structure--property data for rapid mechanical response estimation, and (2) an LLM-based evaluator that jointly considers the design prompt, the candidate structure, and the predicted properties to assess semantic alignment and propose refined feedback. In this way, Supervisor does not replace high-fidelity simulation entirely; instead, it reduces reliance on expensive simulation during the inner optimization loop and improves coordination between language guidance, geometric generation, and property awareness.

Formally, let $\mathcal{V}$ denote the language space, where the human prompt is $V_p \in \mathcal{V}$ and the scaffold generated by Designer is $V_m \in \mathcal{V}$. Let $\mathcal{M}$ denote the geometry space, where Generator maps a latent code $\mathbf{z}\in\mathbb{R}^d$ to a generated structure $\overline{M}\in\mathcal{M}$. We use $f_{\mathcal{V} \rightarrow \mathcal{M}}$ and $f_{\mathcal{M} \rightarrow \mathcal{V}}$ to denote the conversions between scaffold descriptions and geometric representations. Mechanical properties are denoted by $\mathbf{y}\in\mathbb{R}^{d_y}$. Given either a scaffold or a generated structure, Supervisor predicts properties and returns an evaluation signal that drives the two collaboration loops. Algorithm~\ref{alg:collaborations} summarizes the Designer--Supervisor collaboration $C_{D/S}$ and the Generator--Supervisor collaboration $C_{G/S}$; more details are provided in Appendix~\ref{app:collab}.

\vspace{-0.7em}
\subsection{Multi-Agent Collaboration Mechanism}
\vspace{-0.7em}
\fname\ coordinates Designer, Generator, and Supervisor to form a closed-loop discovery process across language, geometry, and property modalities (Figure~\ref{fig:framework}). Starting from a human-authored prompt $V_p$, Designer interprets the design intent and retrieves domain knowledge to produce a scaffold $V_m$, which serves as a semantically aligned structural prior. Supervisor then evaluates this scaffold in the $C_{D/S}$ loop by taking $(V_p, V_m)$ as input, predicting approximate properties $\mathbf{y}_s$, and returning a score $s$ together with a refined prompt $V_p'$. This refined prompt is fed back to Designer to update the scaffold until satisfactory scaffold-level alignment is achieved.

Conditioned on the refined scaffold, Generator performs scaffold-guided geometric synthesis. Specifically, it takes $V_m$ together with optional property targets $\mathbf{y}$ and optional initialization structures, and outputs a candidate metamaterial $\hat{\mathcal{M}}$ through the aforementioned generation steps. Supervisor then closes the loop through $C_{G/S}$ by evaluating $(V_p, \hat{\mathcal{M}})$, predicting effective properties $\mathbf{y}_m$, and returning both a score $s$ and refined property feedback $\mathbf{y}_m'$, which can be used to further steer Generator. In this way, Designer refines semantic intent, Generator refines geometric realization, and Supervisor aligns both with desired property trends. The pipeline terminates when the generated design satisfies the evaluation criterion, after which it is sent to downstream validation such as finite-element simulation, homogenization, or fabrication. Although the present work focuses on a fully automated loop, expert intervention can be naturally incorporated in practical deployment.
\vspace{-0.7em}
\section{Experiments}
\vspace{-0.7em}
To evaluate \fname, we investigate four questions:
\textbf{Q1}:~Validity. Does \fname~generate metamaterials that are more structurally valid than existing baselines?
\textbf{Q2}:~Diversity. Does \fname~generate diverse structures that go beyond the existing design space?
\textbf{Q3}:~Language-Guidance Effectiveness. How effectively do natural-language prompts steer the generation process?
\textbf{Q4}:~Operator-based Programmability. Do the proposed symbolic logic operators enable controllable and meaningful generation?
Finally, to further analyze \fname, we ablate the contributions of the latent evolution objectives and study the effectiveness of the disentangled architecture for property prediction.
\begin{table*}[tb]
\setlength{\tabcolsep}{2pt}
\small
\centering
\vspace{-2em}
\begin{tabular}{l|cccc|cc}
\toprule
{Approach} & $\mathcal{V}_{S}$\%$\uparrow$ & $\mathcal{V}_{P}$\%$\uparrow$ & Cov R.\% $\uparrow$ & \makecell[c]{Repeat\\Ratio\%$\downarrow$}  & \makecell[c]{Prompt Guide\\score (GPT-4.1)$\uparrow$} & \makecell[c]{Repeat\\Num.$\downarrow$} \\
\midrule
\multicolumn{7}{c}{\cellcolor{blue!5}{\textit{Generative Models}}}\\
CDVAE~\citep{CDVAE}            & 57.03 & 0.40  & 55.85      & N/A  & N/A  & N/A\\
DiffCSP~\citep{diffcsp}        & 34.46 & 6.50  & 95.80      & N/A  & N/A  & N/A\\
SyMat~\citep{SyMat}            & 41.10 & 0.00  & 79.34      & N/A  & N/A  & N/A\\
Cond‑CDVAE~\cite{condCDVAE}   & 19.37 & 2.00  & 68.60      & N/A  & N/A  & N/A\\
\multicolumn{7}{c}{\cellcolor{blue!5}{\textit{LLMs}}}\\
GPT-4o-mini~\citep{gpt-4o}                   & 47.29 & 0.00   & 73.6   & 39.66   & 0.4155  & 59 \\
Llama-4-maverick~\citep{touvron2023llama}    & 32.06 & 0.82  & 65.1   & 93.39   & 0.4463 & 80 \\
Qwen3-235b~\citep{qwen2.5}                   & 41.22 & 4.95  & 97.1   & 83.47   & 0.3820  & 73  \\
Deepseek-chat~\citep{liu2024deepseek}        & 46.90 & 16.53 & 73.1   & 85.95   & 0.4189  & 65  \\
Gemini-2.0-flash-lite~\citep{gemini}         & 44.31 & 41.86 & 27.5   & 92.56   & 0.4755  & 67  \\
Deepseek-Reasoning~\cite{guo2025deepseek} & \cellcolor{gray!10}85.5 & 65.30 & 86.9   & 67.7   & 0.4993  & 76 \\
\midrule
\fname~\scriptsize{(Ablation with Pure LLM Agents)}& 78.00 & 29.35 & 88.9   & 51.20   & 0.4832  & 61 \\
\fname~\scriptsize{(Gemini2.0, Mix)}   & 64.53 & 91.74   & 93.3  & \cellcolor{gray!30}0.00   & \cellcolor{gray!20}0.5464 & \cellcolor{gray!30}0      \\
\fname~\scriptsize{(GPT-4o-mini, Mix)}      & 76.84 & \cellcolor{gray!10}94.17   & \cellcolor{gray!10}98.2  & \cellcolor{gray!20}0.83   & \cellcolor{gray!10}0.5234  & \cellcolor{gray!30}0\\
\fname~\scriptsize{(Gemini2.0, Union)} & \cellcolor{gray!20}89.65 & \cellcolor{gray!20}95.97  & \cellcolor{gray!30}99.2   & 10.07  & 0.4966 & 56\\
\fname~\scriptsize{(GPT-4o-mini, Union)}    & \cellcolor{gray!30}91.31 & \cellcolor{gray!30}98.35  & \cellcolor{gray!10}98.7   & \cellcolor{gray!10}7.43   & \cellcolor{gray!30}0.5531  & \cellcolor{gray!20}40\\
\bottomrule
\end{tabular}
\vspace{-0.5em}
\caption{Quantitative comparisons. Validity metrics include symmetry ($\mathcal{V}_{S}$\%) and periodicity ($\mathcal{V}_{P}$\%). Diversity metrics include coverage recall (Cov R.\%), and Repeat Ratio. The prompt guidance metric is denoted as Prompt Guide Score (averaging three runs). 
}
\vspace{-1em}
\label{tab:quant_comp}
\end{table*}
\subsection{Quantitative Comparison}\label{sec:quantitative}
\begin{figure*}[tb]
    \centering
    \vspace{-2em}    
    \includegraphics[width=\linewidth]{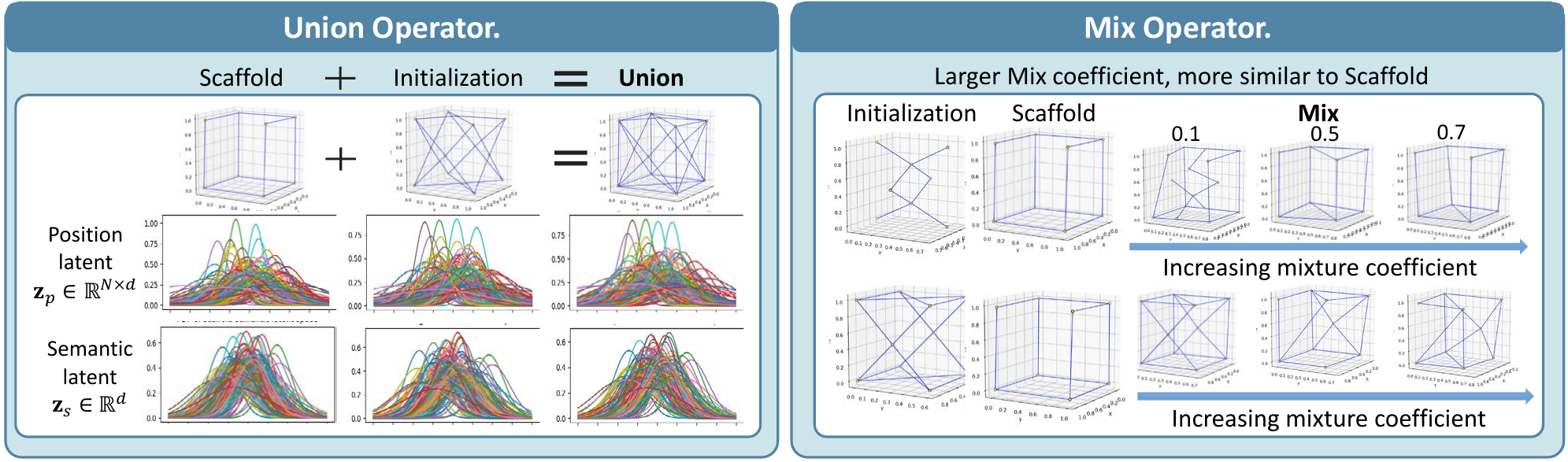}
    \vspace{-1.5em}
    \caption{Qualitative analysis shows the proposed operators and latent evolution methods can successfully program the initialized structure towards the semantic of scaffold.}
    \label{fig:all_operators}
    \vspace{-1em}
\end{figure*}
\paragraph{Experimental Setup.}
We compare \fname~against four geometry-aware generative models and six LLM baselines, including two reasoning-oriented LLMs. Experiments are conducted on the MetaModulus dataset~\citep{Modulus,chen2025metamatbench} using an 8/2 train/test split, together with 100 metamaterial design prompts from domain experts for prompt-based evaluation. We evaluate performance from three perspectives: \textit{(1) Validity}, measured by symmetry ($\mathcal{V}_S$) and periodicity ($\mathcal{V}_P$); \textit{(2) Diversity}, measured by Coverage Recall (Cov.~R.) and Repeat Ratio; and \textit{(3) Language-guidance effectiveness}, measured by a prompt-guidance score produced by an external Supervisor built from GPT-4.1 and a full-data-trained predictor. In the case study, we further validate selected designs via 3D printing as wet-lab testing. More details are provided in Appendix~\ref{app:exp_set}.
\vspace{-1em}
\paragraph{Validity.}
\fname~with the Union operator achieves the strongest validity on both symmetry ($\mathcal{V}_S$) and periodicity ($\mathcal{V}_P$). In particular, \fname~(GPT-4o-mini, Union) reaches 91.31\% on $\mathcal{V}_S$ and 98.35\% on $\mathcal{V}_P$, while \fname~(Gemini-2.0, Union) follows closely with 89.65\% and 95.97\%, respectively. By comparison, the strongest generative baseline, CDVAE, attains only 57.03\% on $\mathcal{V}_S$ and 0.40\% on $\mathcal{V}_P$, while the strongest reasoning LLM baseline, DeepSeek-R, reaches 85.5\% and 65.3\%. These results answer \textbf{Q1} by showing that \fname~substantially improves structural validity over both generative and LLM-based baselines.
\vspace{-1em}
\paragraph{Diversity.}
All \fname~variants achieve high Coverage Recall (\(>90\%\)) on the test set while maintaining low Repeat Ratio (\(<11\%\)). This suggests that \fname~can discover diverse metamaterials that go beyond the training design space rather than merely reproducing seen patterns. In contrast, LLM baselines exhibit substantially higher repetition and generally lower coverage. Notably, GPT-4o-mini attains a relatively low Repeat Ratio of 39.66\%, yet completely fails on periodicity (\(0\%\ \mathcal{V}_P\)), indicating that low repetition alone does not imply meaningful diversity. Overall, these results provide strong evidence for \textbf{Q2}.
\vspace{-1em}
\paragraph{Language-Guidance Effectiveness.}
We measure prompt alignment using the prompt-guidance score from the external Supervisor. \fname~(GPT-4o-mini, Union) achieves the highest score of 0.5531, followed by \fname~(Gemini-2.0, Mix) at 0.5464. All \fname~variants outperform the LLM baselines, whose scores range from 0.3820 to 0.4993, indicating stronger controllability and better alignment with the intended design semantics. These results answer \textbf{Q3} by showing that \fname~more effectively translates natural-language intents into geometric designs.
\vspace{-0.5em}
\subsection{Model Analysis}
\vspace{-0.5em}
\paragraph{Qualitative Analysis.}\label{sec:quali}
We conduct qualitative analysis to answer \textbf{Q4}, focusing on whether symbolic-driven latent evolution enables controllable semantic composition. Figure~\ref{fig:all_operators} presents the qualitative analysis of Union and Mix. More details about Intersection and Negation can be found in Appendix~\ref{app:qual}.
As Figure~\ref{fig:all_operators} left shows, Union combines characteristics from both the scaffold and the initialization, which also is reflected in the fused latent distributions in $\mathbf{z}_p$ and $\mathbf{z}_s$. Figure~\ref{fig:all_operators} right shows that, as $\lambda_{\mathrm{mix}}$ increases, the generated structure gradually shifts from the initialization toward the scaffold, demonstrating smooth and controllable interpolation.
\vspace{-1em}
\paragraph{Ablation Study.}\label{sec:ablation}
\begin{wraptable}{r}{0.3\columnwidth}
\vspace{-1em}
\small
\centering
\footnotesize
\setlength{\tabcolsep}{3pt}
\renewcommand{\arraystretch}{0.95}
\begin{tabular}{@{}lccc@{}}
\toprule
Variant & $\mathcal V_S\uparrow$ & $\mathcal V_P\uparrow$ & Cov.\ R$\uparrow$ \\
\midrule
w/o $\mathcal{L}_s$       & \cellcolor{gray!10}50.9 & \cellcolor{gray!10}57.1 & \cellcolor{gray!30}93.1 \\
w/o $\mathcal{L}_{p,e}$   & \cellcolor{gray!30}47.8 & \cellcolor{gray!30}45.7 & \cellcolor{gray!10}93.6 \\
w/o $\mathcal{L}_r$       & 51.6 & 62.8 & 94.3 \\
w/o $\mathcal{L}_{prior}$ & 58.0 & 62.8 & 95.1 \\
\bottomrule
\end{tabular}
\vspace{-0.8em}
\caption{Ablation study.}\label{tab:ablation}
\vspace{-2em}
\end{wraptable}
We first ablate geometry and property-space awareness by replacing Generator and Supervisor with pure LLM agents while preserving the collaboration mechanism. As shown in Table~\ref{tab:quant_comp}, this variant remains highly repetitive and yields low validity, highlighting the necessity of geometry-aware generation and property-space supervision. We further ablate each term in Eq.~\eqref{eq_main:objective} in Table~\ref{tab:ablation}. Among them, $\mathcal{L}_{p,e}$ has relatively large influence, showing the effectiveness of Sinkhorn-weighted node/edge alignment in latent evolution. Removing $\mathcal{L}_{prior}$ or $\mathcal{L}_r$ also noticeably hurts validity, as both help preserve feasible structures during evolution. In contrast, $\mathcal{L}_s$ mainly affects the semantic perturbation and thus influences all metrics.
\vspace{-0.5em}
\paragraph{Property Prediction Performance.}\label{sec:valid_disent}
\begin{wrapfigure}{r}{0.2\linewidth}
\small
\vspace{-2.5em}
\includegraphics[width=\linewidth,height=8em]{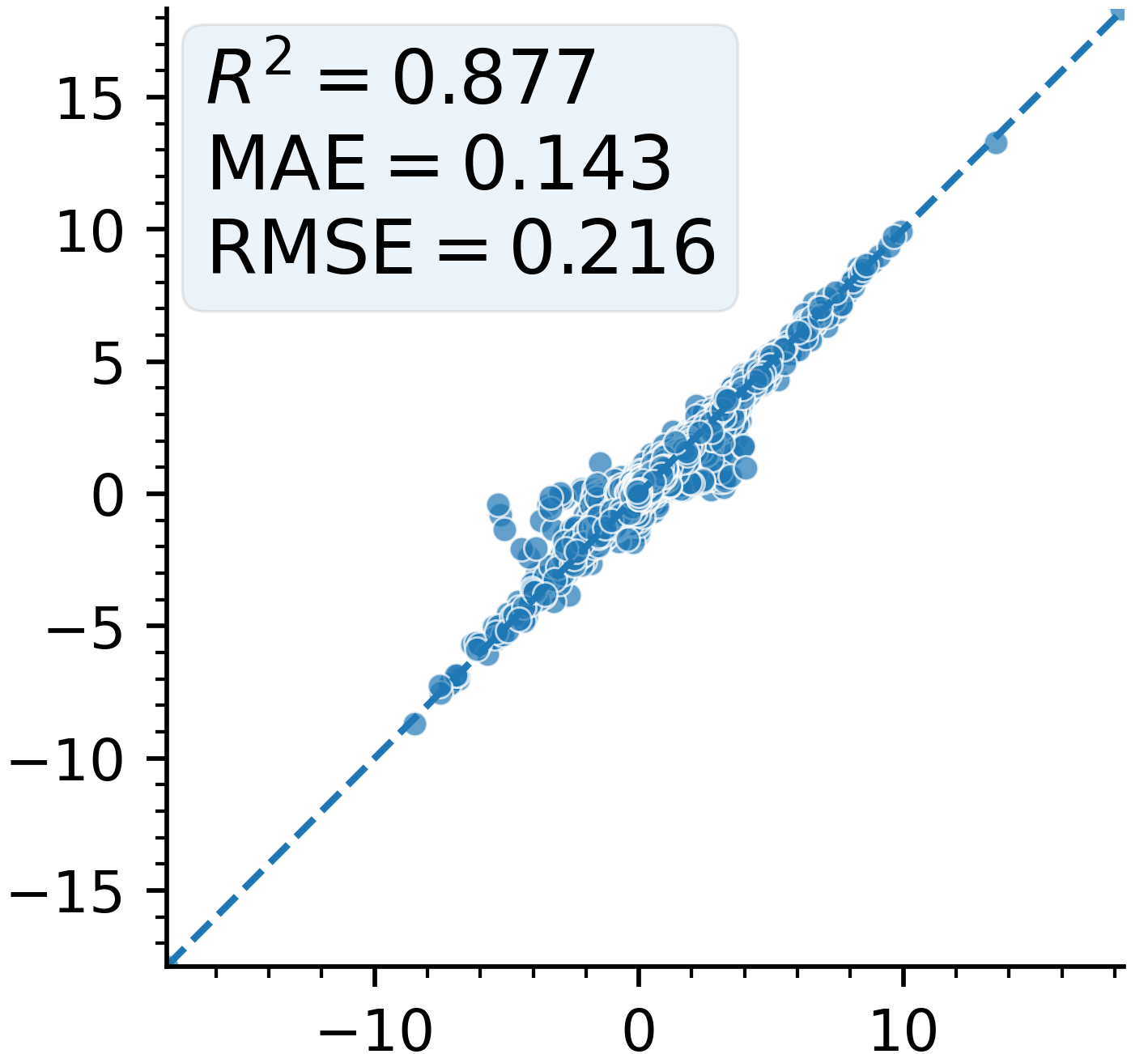}
\vspace{-2em}
\caption{Property prediction.}
\label{fig:prop_pred}
\vspace{-3em}
\end{wrapfigure}
We evaluate the introduced property prediction model used in Supervisor. Figure~\ref{fig:prop_pred} shows the parity plot for Poisson's ratio, where the high $R^2$ of 0.877 indicates strong predictive performance. Additional comparisons with other 3D material representation models are provided in Appendix~\ref{app:more_analy_predictor}, where the proposed architecture also performs superior.
\vspace{-1em}
\begin{figure*}[tb]
\small
\centering
\vspace{-2em}
\includegraphics[width=\linewidth]{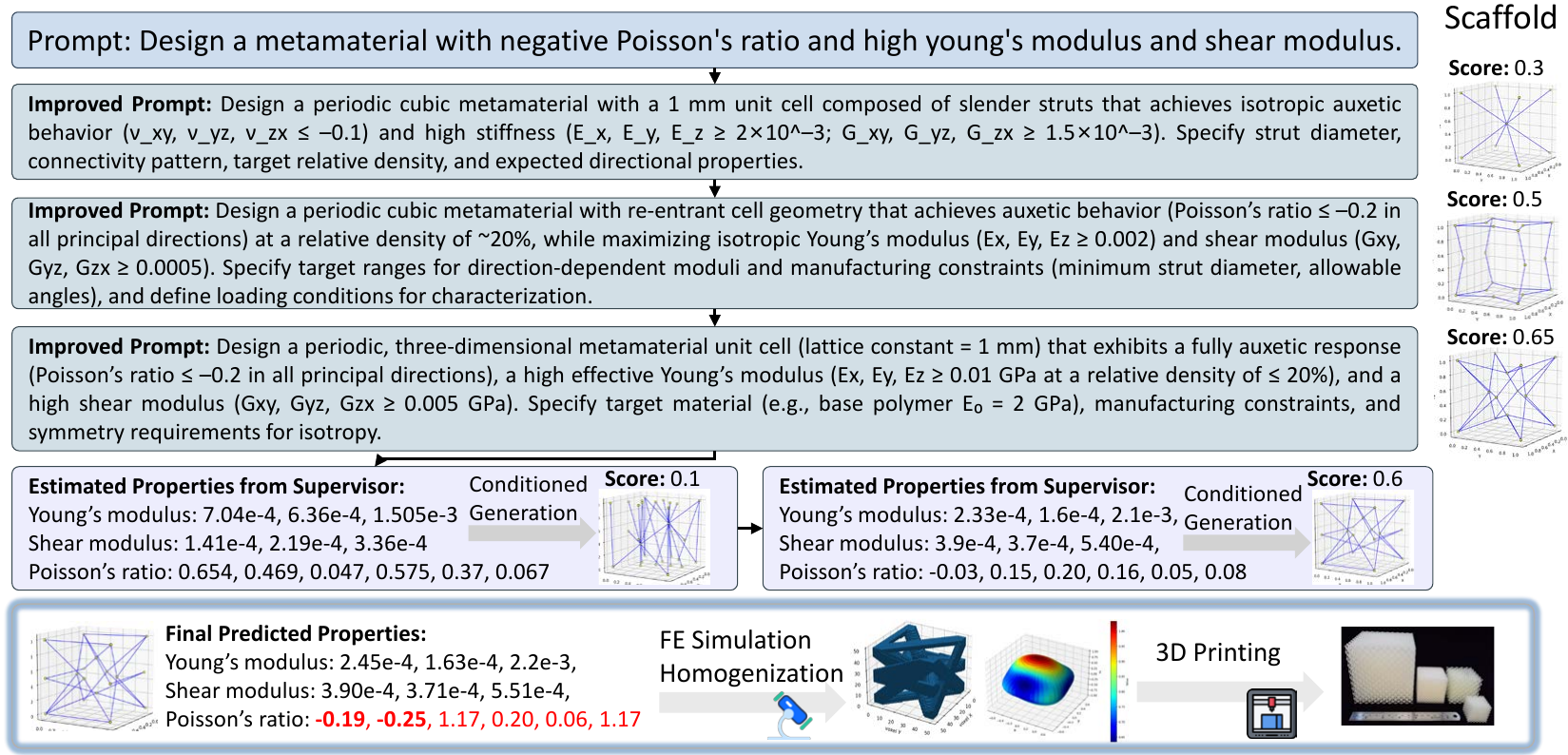}
\vspace{-1em}
\caption{Case study with \fname. FE simulation denotes finite-element simulation.}
\label{fig:case-study}
\vspace{-1em}
\end{figure*}
\section{Case Study}
\vspace{-0.5em}
Figure~\ref{fig:case-study} presents a case study on designing a metamaterial with negative Poisson's ratio and high stiffness using \fname. Starting from a high-level prompt, Designer and Supervisor iteratively refine the design specification by incorporating structural constraints, and target mechanical properties. The final refined prompt achieves a score of 0.65 and specifies a periodic, isotropic structure with auxetic behavior and high modulus. Conditioned on the resulting scaffold, Agent Generator and Agent Supervisor then identify a candidate initialization and optimize it toward improved mechanical properties. The final design is further validated through simulation and asymptotic homogenization~\citep{ANDREASSEN2014488,ARABNEJAD2013249,homogenizationcode}, and subsequently fabricated via 3D printing for experimental verification. The predicted properties of the selected structure include negative Poisson's ratios (\ie, $-0.19$ and $-0.25$) together with relatively high shear modulus terms, demonstrating the practical potential of \fname~for real-world metamaterial discovery.

\vspace{-1em}
\section{Conclusion}
\vspace{-1em}
In this paper, we present \fname, a multi-agent framework for language-guided metamaterial discovery that bridges language, geometry, and property modalities through coordinated reasoning, generation, and evaluation. Its key technical contribution is \textbf{symbolic-driven latent evolution} in a \textbf{disentangled latent space}, which enables controllable inference-time refinement beyond simple retrieval or sampling. Across extensive experiments, \fname\ substantially improves structural validity over strong generative and LLM baselines, while also achieving stronger prompt alignment and maintaining high diversity. The real-world auxetic high-stiffness case study further shows that the framework supports an end-to-end workflow from prompt refinement to practical validation. Overall, our results suggest that language can serve as a practical interface for early-stage metamaterial discovery when coupled with geometric generation and fast property-aware feedback.



\section*{Ethics Statement}
\fname\ presents a novel approach to language-guided metamaterial discovery that could significantly enhance the accessibility and efficiency of advanced materials design. By enabling intuitive, language-based, and human-AI collaborative programming over structural generation and property optimization, this framework facilitates metamaterial design for non-experts and accelerates early-stage innovation in domains such as robotics, aerospace, and biomedical engineering.
However, as with any generative design tool, there are potential risks. Misuse of the technology, for instance, in the automated design of materials for harmful or ethically ambiguous applications, could lead to unintended consequences. To mitigate these risks, responsible deployment should include human-in-the-loop validation, interpretability safeguards, and domain-specific oversight. 
Overall, \fname\ has the potential to positively transform the scientific design workflow while necessitating thoughtful governance in high-impact use cases.

\section*{Statement of LLM Usage}
Large language models (LLMs) were used in this work in three ways: as integral components of our framework (Agent Designer and Agent Supervisor), as baselines for experimental comparison (\eg, GPT-4o-mini, Llama-4-Maverick, Gemini-2.0-flash-lite, Qwen3-235b, Deepseek-chat, Deepseek-Reasoning), and as assistive tools to polish the manuscript’s language and presentation; all scientific ideas, methodological contributions, experimental designs, and final claims were conceived and verified by the authors.

\bibliography{main}
\bibliographystyle{colm2026_conference}

\appendix

\section{Detailed Related Works}
\label{app:related}
\subsection{Metamaterial Inverse Design}\label{app:related_inverse}
The inverse design of metamaterials involves generating microstructures that achieve user-specified mechanical responses, such as target elastic moduli or Poisson's ratios. Traditional approaches rely on topology optimization or evolutionary strategies~\citep{RW006}, but these methods often struggle with multi-objective formulations and are computationally expensive.

To overcome these limitations, ML models have been introduced to learn structure--property relationships and enable efficient inverse mapping~\citep{RW005}. Early works apply variational autoencoders (VAEs)~\citep{RW001, RW003,10833915}, graph neural networks (GNNs)~\citep{customGNN,smhgc}, or generative adversarial networks (GANs) to model the one-to-many nature of the inverse problem, capturing diverse valid solutions. More recent efforts such as CDVAE~\citep{CDVAE}, Cond-CDVAE~\citep{condCDVAE}, and SyMat~\citep{SyMat} extend this paradigm to periodic and symmetric materials, incorporating physical invariance and conditioning on accurate parameters. DiffCSP~\citep{diffcsp} furthers this by applying diffusion models over periodic structures with equivariant geometry.

Beyond generative modeling, approaches like Deep-DRAM~\citep{RW003} and Cycle-GAN~\citep{RW001} offer modular solutions for multi-objective or deformation-dependent inverse design. These frameworks demonstrate size-agnostic predictions, property-aware sampling, and resistance to fatigue or fracture. However, they typically rely on numerical conditioning and do not accept conceptual or language design queries.

Our work addresses this gap by enabling language-conditioned inverse design through multi-agent collaborations. Compared to prior methods that require exact target vectors, our framework supports symbolic and language prompts and incorporates multi-modality agents that operate over geometry, physics properties, and language. This design philosophy enables more interpretable, flexible, and interactive design workflows.

\subsection{Agentic AI for Material Discovery}\label{app:related_agentic}
Recent advances in LLMs have shown remarkable potential in augmenting scientific and engineering workflows through reasoning, planning, and symbolic manipulation. In the metamaterial domain, where design spaces are combinatorially vast and high-fidelity evaluation is computationally expensive, LLMs offer a promising interface for intuitive, high-level control of generative pipelines.

Several works integrate LLMs with physical constraints for 3D material generation. For example, ESM3~\cite{esm3} jointly models sequence, structure, and function via tokenized multimodal prompts; BindGPT~\cite{bindgpt} generates 3D molecules via a language model trained on spatial data;  CrystaLLM~\citep{CrystaLLM} fine-tunes LLMs to generate inorganic crystal structures as text; EquiLLM~~\citep{EquiLLM} combines LLMs with equivariant GNNs to model physical systems; Geo2Seq~\citep{Geo2Seq} introduces SE(3)-invariant tokenization for LLMs to generate 3D molecules. However, these works based on tokenization of geometric structures, have limited exploration in geometric space. MetaSymbO similarly spans modalities, but uniquely enables inference-time symbolic composition and extrapolative design beyond training distributions.

In addition, the latest related works, such as CrossMatAgent~\citep{tian2025multi}, demonstrate multi-agent systems that couple LLMs with visual generation models and physics-based simulators to automate design tasks. These systems enable LLMs to act as supervisory agents that propose structures, query simulations, and refine designs, leveraging capabilities from models like GPT-4o~\citep{gpt-4o}, DeepSeek~\citep{liu2024deepseek}, Gemini~\citep{gemini}, and Qwen2.5~\citep{qwen2.5} for multimodal understanding and structured reasoning.

Complementary research~\citep{RW101} frames LLMs as autonomous mechanical designers capable of iteratively generating and refining truss structures via feedback from finite element analysis (FEA), achieving performance competitive with traditional optimization methods. These findings support the feasibility of deploying LLMs in complex inverse design settings without task-specific training.

However, challenges remain: LLMs often lack geometric awareness and are not pretrained on physical design tasks. To bridge this gap, our approach introduces LLMs as language agents operating in a cooperative multi-agent setting. Rather than relying on zero-shot generation alone, our system aligns language-derived intents with latent structural priors and property constraints, amplifying LLM utility via multimodal feedback loops.

By combining the natural abstraction power of LLMs with geometry-aware agents and predictive supervision, we demonstrate that LLMs can serve not only as query interfaces but as active participants in the multi-agent collaboration framework.

\section{Implementation Details}\label{app:imple}
\subsection{Details of Prompt for Agents Instantiation}\label{app:prompt_details}
\begin{figure}[h]
    \centering
    \includegraphics[width=\linewidth]{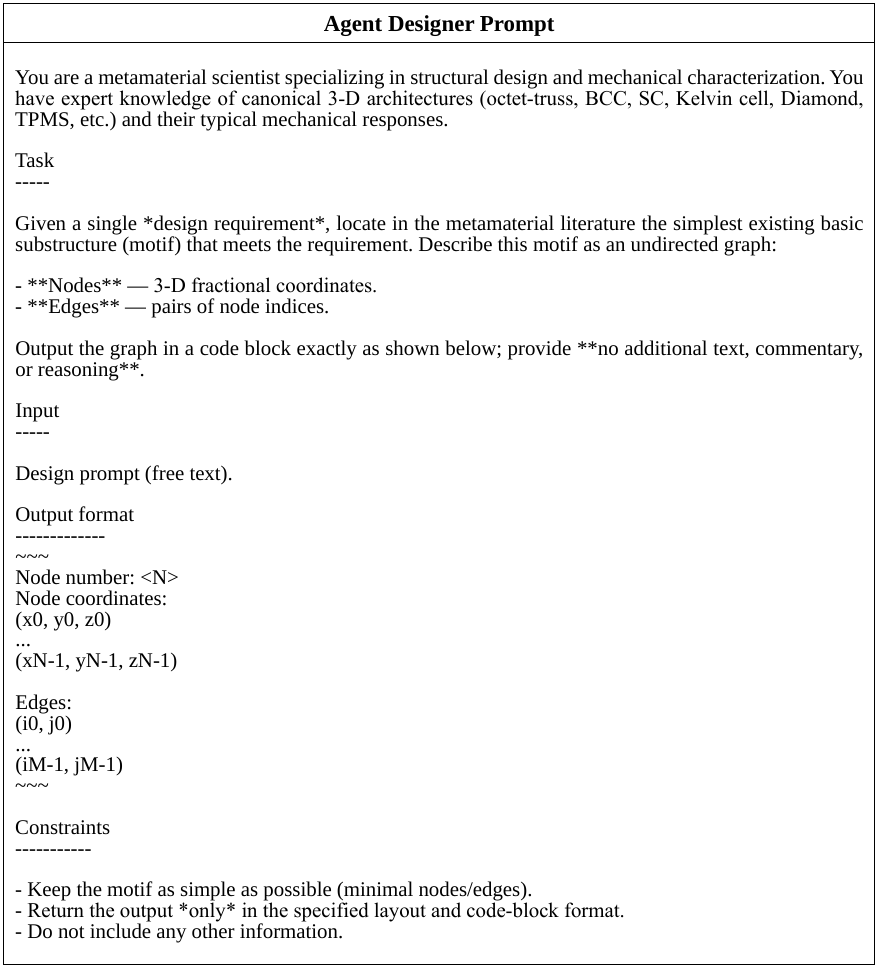}
    \caption{Prompt for Designer.}
    \label{fig:prompt_design}
\end{figure}

\begin{figure}[h]
    \centering
    \includegraphics[width=\linewidth]{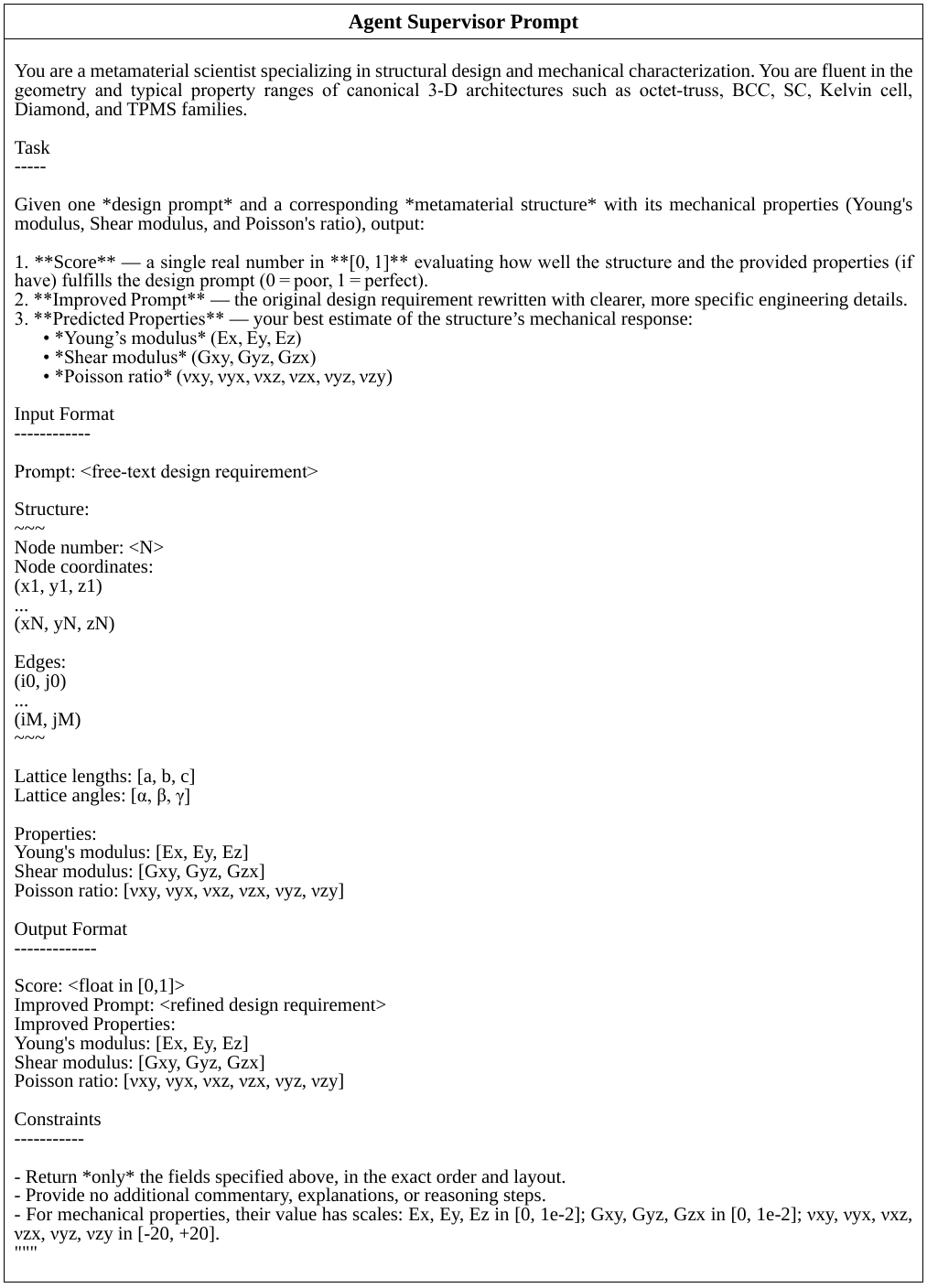}
    \caption{Prompt for Supervisor.}
    \label{fig:prompt_supervisor}
\end{figure}

To instantiate the goals as we discussed in Section~\ref{sec:agents_intro}, we provide the detailed prompts for Agent Designer and Agent Supervisor to deal with the inputs, outputs, and constraints.
Figures~\ref{fig:prompt_design} and~\ref{fig:prompt_supervisor} illustrate the prompt for the designer and supervisor. Specifically, Designer highlights \textit{locate} from existing literature for a simple structure containing semantics similar to the input text. Alternatively, the supervisor emphasizes utilizing not only the literature but also the predicted mechanical properties to obtain the score, as well as an improved prompt that will provide feedback to designer for the next design iteration.

\clearpage

\subsection{Details of Collaboration Mechanisms with Supervisor}\label{app:collab}
Algorithm~\ref{alg:collaborations} shows the collaboration steps of Agent 1\&3 collaboration and Agent 2\&3 collaboration. To clarify, we further introduce the collaboration steps as follows in detail.
\paragraph{Agent Designer\&Supervisor Collaboration.} Agent Designer receives the prompt $V_p$ from human and outputs the best-matched basic structures (\ie, scaffold) $V_m=A_1(V_p)$ by exploiting its literature base. After that, the scaffold is fed into Agent Supervisor, which first predicts the associated mechanical properties $\mathbf{y}_s = A_3^{\text{pred}}(V_m)$. Based on both the predicted properties $\mathbf{y}_s$ and the original input-output pairs ($V_p$, $V_m$), Agent Supervisor computes a match score and provides an improved prompt $s, V_p'  = A_3^{\text{eval}}(V_m, V_p, \mathbf{y}_s)$. Finally, it returns score $s$ and improved prompt $V_p'$ to Agent Designer. This loop repeats till a good evaluation score is received. Formally, using $t$ to denote iteration and $\tau_{D/S}$ be threshold.

\paragraph{Agent Generator\&Supervisor Collaboration.} Agent Generator begins by initializing a latent Gaussian noise vector or randomly selects an initialization from dataset, which serves as the starting point in the generation process, denoted as $\overline{M} = A_2(\mathbf{z})$, where $\mathbf{z} \sim \mathcal{N}(\mathbf{0}, \mathbf{I})$. The generated structure $\overline{M}$ is then passed to Agent Supervisor to predict its mechanical properties as $\mathbf{y}_m = A_3^{\text{pred}}(\overline{M})$. Subsequently, Agent Supervisor evaluates the quality of the structure by computing a match score $s$ and generating an updated property target $\mathbf{y}_m'$ using the evaluation function $A_3^{\text{eval}}$. The updated properties $\mathbf{y}_m'$ are then (optionally) used to guide the initialization of the latent vector $\mathbf{z}$ for the next iteration in Agent Generator. This iterative process continues until the evaluation score $s$ reaches the threshold $\tau_{G/S}$ as shown in the right column of Alg.~\ref{alg:collaborations}.

\begin{figure}[h]
    \centering
    \includegraphics[width=0.9\linewidth]{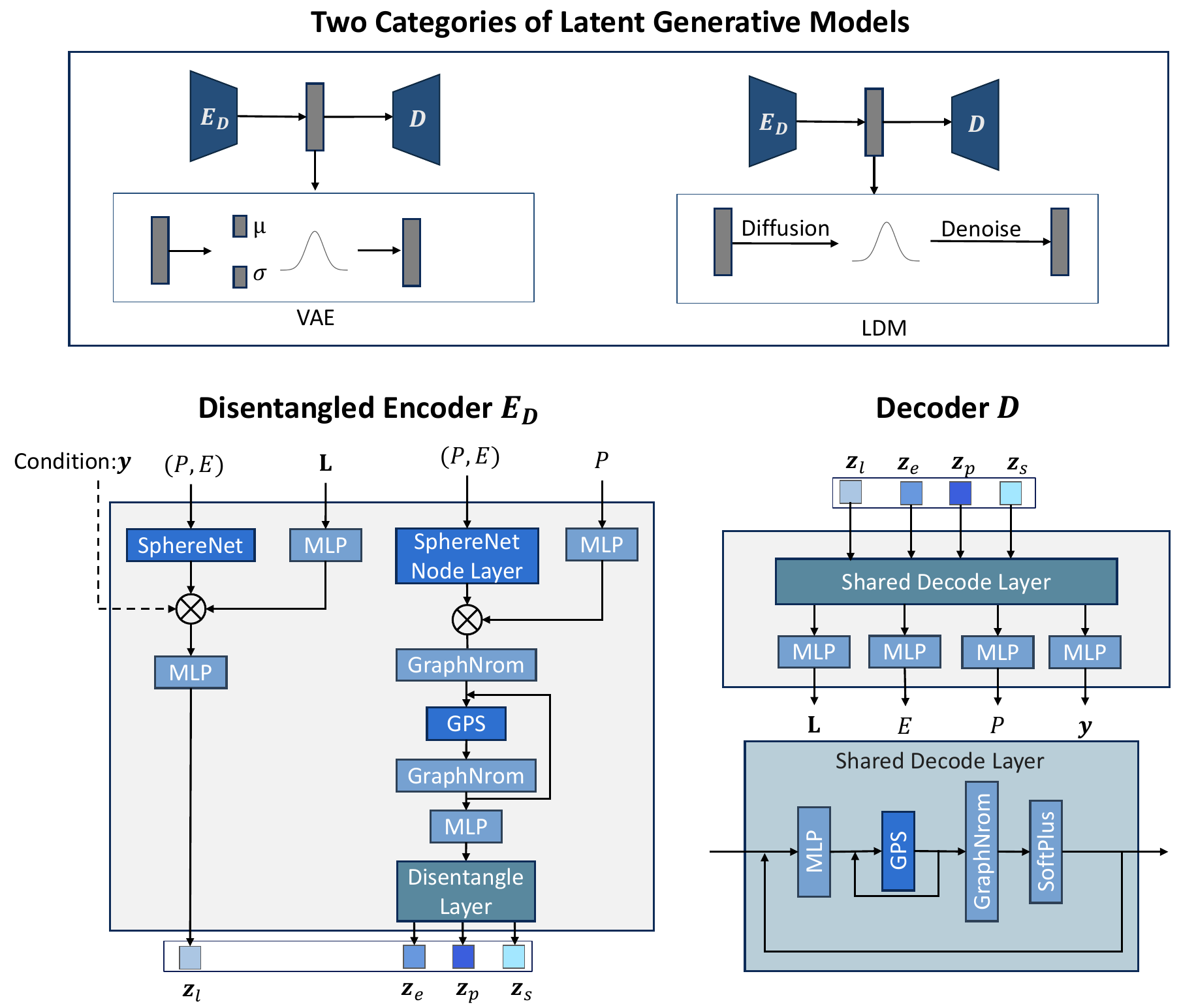}
    \caption{Implementation details of the disentangled encoder and decoder. After the geometry is encoded to a disentangled latent Gaussian space, it is trivial to implement VAEs and DMs for generation as the upper part shown. Please find more details in our codebase.}
    \label{fig:enc_dec}
\end{figure}
\subsection{Details of Symbolic-Driven Latent Evolution}
\label{app:symbo_opt}
In this section, we detail the process in Generator, especially the disentangled generation and symbolic operators.

\paragraph{Disentangling Latent Generation.} Figure~\ref{fig:enc_dec} describes the implementation of the encoder and decoder for latent space construction. Considering the complete metamaterial representations $\mathcal{M} = (\mathbf{L}, \mathcal{U})$ and $\mathcal{U}=(\mathbf{P}, E)$, with four representation dimension, \ie, lattice vector $\mathbf{L}$, associated property $\mathbf{y}$, node positions $\mathbf{P}$ and edges $E$, we disentangle the latent $\mathbf{z}$ to $\mathbf{z}_l, \mathbf{z}_p, \mathbf{z}_e, \mathbf{z}_s$ implying the latice, coordinate position, edges, and metamaterial semantics (referring to the mechanical properties) with Gaussian prior:
\begin{equation}\label{eq:enc}
\small{
    \begin{aligned}
    &q_\phi(\mathbf{z} \mid M) = q_\phi(\mathbf{z}_l, \mathbf{z}_p, \mathbf{z}_e, \mathbf{z}_s \mid M) = 
    q_{\phi_1}(\mathbf{z}_l \mid M)\, q_{\phi_2}(\mathbf{z}_p \mid M)\, q_{\phi_3}(\mathbf{z}_e \mid M)\, q_{\phi_4}(\mathbf{z}_s \mid M), \\
    &p(\mathbf{z}) = p(\mathbf{z}_l)p(\mathbf{z}_p) p(\mathbf{z}_e)p(\mathbf{z}_s) \text{, where each } p(\cdot) \sim \mathcal{N}(\mathbf{0}, \mathbf{I}).
\end{aligned}
}
\end{equation}

Therefore, the metamaterial-specific decoder reconstruct the full metamaterial spaces $\mathcal{M}$ using conditional likelihood:
\begin{equation}\label{eq:dec}
\small{
    p_{\theta}(M\mid \mathbf{z}) = p_{\theta_1}(\mathbf{L}\mid \mathbf{z_1}) p_{\theta_2}(\mathbf{P} \mid \mathbf{z_p}) p_{\theta_3}(E \mid \mathbf{z_e}) p_{\theta_4}(\mathbf{y}\mid \mathbf{z_s}),
}
\end{equation}

which imposes the four latent spaces with specific physical meanings.
Finally, the derived ELBO for the disentangled VAE can be expressed as:
\begin{equation}\label{eq:elbo}
\small{
\begin{aligned}
\mathcal{L}_{\text{ELBO}}(\theta, \phi; \mathcal{M}) 
= &\; \mathbb{E}_{q_\phi(\mathbf{z} \mid \mathcal{M})} \left[
\log p_{\theta_1}(\mathbf{L} \mid \mathbf{z}_l)
+ \log p_{\theta_2}(\mathbf{P} \mid \mathbf{z}_p)
+ \log p_{\theta_3}(E \mid \mathbf{z}_e)
+ \log p_{\theta_4}(\mathbf{y} \mid \mathbf{z}_s)
\right] \\
&- \mathrm{KL}\left(q_{\phi_1}(\mathbf{z}_l \mid \mathcal{M}) \,\|\, p(\mathbf{z}_l)\right)
- \mathrm{KL}\left(q_{\phi_2}(\mathbf{z}_p \mid \mathcal{M}) \,\|\, p(\mathbf{z}_p)\right) \\
&- \mathrm{KL}\left(q_{\phi_3}(\mathbf{z}_e \mid \mathcal{M}) \,\|\, p(\mathbf{z}_e)\right)
- \mathrm{KL}\left(q_{\phi_4}(\mathbf{z}_s \mid \mathcal{M}) \,\|\, p(\mathbf{z}_s)\right)
\end{aligned}
}
\end{equation}

Similarly, we can derive the objective for disentangled LDM with the AE (Eqs.~\ref{eq:enc} and \ref{eq:dec}) as follows: 
\begin{equation}\label{eq:score}
\small{
    \mathcal{L}_{\text{LDM}}^{(t)}(\theta) =
\sum_{i \in \{l, p, e, s\}} 
\mathbb{E}_{\mathbf{z}_i^{(0)}, \boldsymbol{\epsilon}_i, t}
\left[
\left\|
\boldsymbol{\epsilon}_i - \boldsymbol{\epsilon}_{\theta_i}(\mathbf{z}_i^{(t)}, t)
\right\|^2
\right].
}
\end{equation}
Note that the four encoder parameters $\{\phi_i\}^4_{i=1}$ and decoder/denoising parameters $\{\theta_i\}^4_{i=1}$ contain shared layers, and only the final head layers need to be fine-tuned for disentangling. In addition, we can implement conditional generation by concatenating condition $\mathbf{y}_{cond}$ in the decoder/denoising model as $p_{\theta}(M\mid \mathbf{z}, \mathbf{y}_{cond})$ or $\epsilon_\theta(\mathbf{z}, t, \mathbf{y}_{cond})$.

With this design, we are able to control the generation process regarding all four aspects of a metamaterial in four latent spaces, respectively, achieving a fine-grained generation control.

\paragraph{Symbolic Logic Operators.} We introduce four symbolic logic operators for the scaffold guidance, \ie, Union, Mix, Intersection, and Negation.

Union aims to expand the node set of the source metamaterial $M$ according to the guidance scaffold $M'$ in node level. More than a simple expansion of node set, it further fuses the semantics in semantic space. Formally, let $\mathbf{z}_i\sim \mathcal{N}(\boldsymbol{\mu}_i, \mathbf{\Sigma}_i)$, $i\in \mathcal{I}_M=\{1,...,N_M\}$ and $\mathbf{z}'_j\sim \mathcal{N}(\boldsymbol{\mu}'_j, \mathbf{\Sigma}'_j)$, $j\in \mathcal{I}_M'=\{1,...,N_{M'}\}$ be the node-level latents of original $M$ and scaffold $M'$, where $N_{M}$ and $N_{M'}$ denotes their node numbers. We introduce a differentiable soft matching algorithm--Sinkhorn transport~\citep{frogner2015learning} that measures the probability mass of two distributions being regarded as the same physical node--for differentiable assignment matrix (where differentiability enables gradient-based latent optimization):
\begin{equation}\label{eq:sinkhorn}
    \mathbf{P} = [\mathbf{P}_{ij}]_{i\in\mathcal{I_M}, j \in \mathcal{I}_{M'}} \in [0,1]^{N_M \times N_{M'}}.
\end{equation}
Therefore, $r_i=\sum_j\mathbf{P}_{ij}, c_j = \sum_i\mathbf{P}_{ij}$, $0\le r_i, c_j \le 1$ gives the probability of overlaped nodes in $M$ and $M'$, respectively, and $\rho=\sum_i r_i = \sum_j c_j$ representing the number of overlapping nodes. 
The detailed computation of the Sinkhorn matrix $\mathbf{P}$ is illustrated in Algorithm~\ref{alg:sinkhorn_log}.
\begin{algorithm}[t]
\caption{\textsc{Sinkhorn\_Log}: Log-Stabilised Sinkhorn Iteration}
\label{alg:sinkhorn_log}
\begin{algorithmic}[1]
\REQUIRE Cost matrix $C\in\mathbb{R}^{N_s\times N_t}$, entropic weight $\varepsilon>0$, maximum iterations $T$, tolerance $\tau$
\ENSURE Transport plan $P\in\mathbb{R}^{N_s\times N_t}$

\STATE $K \gets -\dfrac{C}{\varepsilon}$ \algcomment{kernel in the log domain}
\STATE $f \gets \mathbf{0}_{N_s}$;\;\; $g \gets \mathbf{0}_{N_t}$

\FOR{$t = 1,2,\ldots,T$}
    \STATE $f_{\text{prev}} \gets f$

    \STATE \textit{/* row scaling update */}
    \STATE $\tilde{K} \gets K + g^{\top}$ \algcomment{$\tilde{K}_{ij}=K_{ij}+g_j$}
    \STATE $\tilde{K} \gets \tilde{K} - \max_{j}\tilde{K}_{ij}$ \algcomment{row-wise stabilisation}
    \STATE $f \gets -\operatorname{LSE}_{j}(\tilde{K}_{ij})$

    \STATE \textit{/* column scaling update */}
    \STATE $\hat{K} \gets K + f$ \algcomment{$\hat{K}_{ij}=K_{ij}+f_i$}
    \STATE $\hat{K} \gets \hat{K} - \max_{i}\hat{K}_{ij}$ \algcomment{column-wise stabilisation}
    \STATE $g \gets -\operatorname{LSE}_{i}(\hat{K}_{ij})$

    \STATE $\delta \gets \lVert f - f_{\text{prev}} \rVert_{\infty}$
    \IF{$\delta > 10^{4}$}
        \STATE \textbf{break} \algcomment{numerical blow-up guard}
    \ENDIF
    \IF{$\delta < \tau$}
        \STATE \textbf{break} \algcomment{convergence reached}
    \ENDIF
\ENDFOR

\STATE $P \gets \exp\!\bigl(K + f + g^{\top}\bigr)$
\STATE \textbf{return} $P$
\end{algorithmic}
\end{algorithm}

Using $\delta_{(\boldsymbol{\mu}, \mathbf{\Sigma})}$ denotes Dirac measure centered at the parameter pair $(\boldsymbol{\mu},\mathbf{\Sigma})$, which indicates one node in the union has distribution $\mathcal{N}(\boldsymbol{\mu}, \boldsymbol{\Sigma})$, we can construct the discrete measure of Union on continuous Gaussian space as ${\pi}_{\cup}$ in Eq.~\eqref{eq:union} for gradient-based optimization.
\begin{equation} \label{eq:union}
\boxed{
\begin{aligned}
    &{\pi}_{\cup} = \frac{{\mu_{\cup}}}{Z}, \text{ where } Z=N_M + N_{M'} - \rho,  \text{ and }\\
    &\mu_{\!\cup}
    =
    \underbrace{\sum_{i\in\mathcal I_M} (1 - r_i)\,
                \delta_{(\boldsymbol\mu_i, \boldsymbol\Sigma_i)}}_{\text{nodes unique to } M}
    \;+\;
    \underbrace{\sum_{j\in\mathcal I_{M'}} (1 - c_j)\,
                \delta_{(\boldsymbol\mu'_j, \boldsymbol\Sigma'_j)}}_{\text{nodes unique to } M'}
    \;+\;
    \underbrace{\sum_{i,j} P_{ij}\,
                \delta_{(\boldsymbol\mu_i, \boldsymbol\Sigma_i)}}_{\text{overlapping nodes, preserving $M$}}
\end{aligned}
}
\end{equation}

\textit{Mix.} Unlike Union, the Mix operator blends the latent distributions of the original metamaterial $M$ and the scaffold $M'$ into a single composite distribution, where the contribution of the scaffold is modulated by a guidance coefficient $\lambda_{\mathrm{mix}} \in [0,1]$. Its probabilistic form is expressed as:
\begin{equation}
p_{\text{mix}}(\mathbf{z}\mid\lambda_{\text{mix}})
=(1-\lambda_{\text{mix}})p_M(\mathbf{z})
+\lambda_{\text{mix}}p_{M'}(\mathbf{z}),
\label{eq:mixture_prob}
\end{equation}
where $p_M$ and $p_{M'}$ denote the empirical latent distributions induced by $M$ and $M'$ respectively. However, Eq.~\eqref{eq:mixture_prob} is typically intractable due to the complexity of $p_M$ and $p_{M'}$. Considering that latent distributions are Gaussian, we adopt a simplified moment-matching approximation~\citep{bishop2006pattern}:
\begin{equation} \label{eq:mixture}
\small{
\boxed{
\begin{aligned}
p_{\mathrm{mix}}(\mathbf{z}\mid \lambda_{\mathrm{mix}})
\;\approx\;
\mathcal{N}\Bigl(\mathbf{z}\;;\;(1-\lambda_{\mathrm{mix}})\,\boldsymbol{\mu}_{M} + \lambda_{\mathrm{mix}}\,\boldsymbol{\mu}_{M'}
\;,\;\mathrm{diag}\!\bigl((\, (1-\lambda_{\mathrm{mix}})\,\boldsymbol{\sigma}_{M} + \lambda_{\mathrm{mix}}\,\boldsymbol{\sigma}_{M'}\,)^{2}\bigr)\Bigr)
\end{aligned}
}
}
\end{equation}

Intersection operator aims at identifying the common semantics or overlapping components between the two distributions of $M$ and $M'$ in the latent space. To do so, we introduce Product-of-Expert (PoE), which results in the distribution focusing on regions of high probability shared by both $p_M$ and $p_{M'}$. This can effectively sharpen the distribution, making generated samples focus more on shared structure features. Formally, the Intersection of two distribution using PoE is:
\begin{equation}\label{eq:prod}
    p_{\mathrm{int}}(\mathbf{z}) \propto p_{M}(\mathbf{z}) \cdot p_{M'}(\mathbf{z}).
\end{equation}
Considering both $p_{M}(\mathbf{z})$ and $p_{M'}(\mathbf{z})$ hold Gaussian prior, the resulting Intersection distribution can be derived as:
\begin{equation}\label{eq:intersection}
\small{
\boxed{
\begin{aligned}
&p_{\mathrm{int}}(\mathbf{z}) = \mathcal{N}(\mathbf{z}; \boldsymbol{\mu}_{\mathrm{int}}, \boldsymbol{\Sigma}_{\mathrm{int}}),\\
&\text{where } \boldsymbol{\Sigma}_{\mathrm{int}} = (\boldsymbol{\Sigma}_M^{-1} + \boldsymbol{\Sigma}_{M'}^{-1})^{-1}, \text{ and } \boldsymbol{\mu}_{\mathrm{int}} = \boldsymbol{\Sigma}_{\mathrm{int}}(\boldsymbol{\Sigma}_M^{-1}\boldsymbol{\mu}_M + \boldsymbol{\Sigma}_{M’}^{-1}\boldsymbol{\mu}_{M’})
\end{aligned}
}
}
\end{equation}
Negation is contrary to Intersection that emphasizes common high-density regions, Negation aims to suppress the influence of high-density regions in the latent space of $M'$ from that of $M$. Accordingly, we can define the unnormalized probability density as:
\begin{equation}\label{eq:neg_prob}
    p_{\mathrm{neg}}(\mathbf{z}) \propto \frac{p_{M}(\mathbf{z})^{\alpha}}{p_{M'}(\mathbf{z})^\beta},
\end{equation}
where $\alpha, \beta > 0$ are hyperparameters that respectively control the strength of preservation and suppression.

However, under this construction, the resulting distribution $p_{\mathrm{neg}}(\mathbf{z})$ is no longer strictly Gaussian, potentially leading collapse of decoding. Therefore, we perform moment-matching to approximate it as a single Gaussian, as shown in Eq.~\eqref{eq:neg}.
\begin{equation}\label{eq:neg}
\small{
\boxed{
\begin{aligned}
&p_{\mathrm{neg}}(\mathbf{z}) \approx \mathcal{N}(\mathbf{z}; \boldsymbol{\mu}_{\mathrm{neg}}, \boldsymbol{\Sigma}_{\mathrm{neg}}),\\
&\text{where } \boldsymbol{\Sigma}_{\text{neg}}^{-1} = \alpha \boldsymbol{\Sigma}_M^{-1} - \beta \boldsymbol{\Sigma}_{M'}^{-1}, \text{and }
\boldsymbol{\mu}_{\text{neg}} = \boldsymbol{\Sigma}_{\text{neg}} 
\left( \alpha \boldsymbol{\Sigma}_M^{-1} \boldsymbol{\mu}_M - \beta \boldsymbol{\Sigma}_{M'}^{-1} \boldsymbol{\mu}_{M'} \right)
\end{aligned}
}
}
\end{equation}
Here, $\alpha$ and $\beta$ are hyperparameters that control the strength of preservation and negation, respectively. A larger $\beta$ increases the degree of suppression exerted by the latent distribution of $M'$.

\paragraph{Gaussian Latent Evolution.}
Although a symbolic–logic operator yields a closed‑form \emph{target} Gaussian, decoding from that distribution directly poses two problems.  
(1) In a disentangled AE the decoder is trained only on the latent manifold induced by the encoder.  Closed‑form operations such as \emph{Mixture}, \emph{Intersection}, or \emph{Negation} can push the target distribution far outside this manifold, so the decoder may produce implausible geometries.  
(2) Symbolic operators act component‑wise and therefore fuse two latents within the same sub‑space; statistical dependencies across the four disentangled sub‑spaces vanish, breaking the compatibility that the decoder relies on.

To resolve both issues, we \emph{optimize the original latent vector} toward the closed‑form target by gradient descent.  During this process, we impose a Sinkhorn‑based soft‑matching loss on node and edge distributions, which preserves cross‑space coherence and keeps the trajectory on the learned manifold. Formally, the latent optimization loss can be the weighted sum of Eq.~\eqref{eq:objective}. In detail, KL between the semantic distribution $\mathcal{L}_s$ is to learn targeted semantics; Sinkhorn weighted KL between the node position/edge distributions $\mathcal{L}_{p,e}$ is to learn node alignment in both edge and node space; regularization $\mathcal{L}_r$ is to preserve original distributions of non-overlapping nodes; and latent prior $\ell_2$ norm $\mathcal{L}_{\mathrm{prior}}$ is to prevent distribution drift.
\begin{equation}\label{eq:objective}
\small{
\begin{aligned}
    &\mathcal{L}_s  = \text{KL}(\mathcal{N}(\boldsymbol{\mu}_s, \boldsymbol{\sigma}_s)|| \mathcal{N}(\boldsymbol{\mu}_s', \boldsymbol{\sigma}_s')) \text{\quad(Graph-level semantic optimization)},\\
    &\mathcal{L}_{p,e} = \sum_{k\in\{p,e\}}\sum_{i=1}^{N_M}\sum_{j=1}^{N_{M'}}\mathbf{P}_{ij} \text{KL}(\mathcal{N}(\boldsymbol{\mu}_{k,i}, \boldsymbol{\sigma}_{k,i})|| \mathcal{N}(\boldsymbol{\mu}_{k,j}', \boldsymbol{\sigma}_{k,j}'))\text{\quad(Node-level pos./edge alignment)},\\
    &\mathcal{L}_{r}=\sum_{k\in\{p,e\}}\sum_{i\in \{i|r_i<\tau_{o}\}}\text{KL}(\mathcal{N}(\boldsymbol{\mu}_{k,i}, \boldsymbol{\sigma}_{k,i})||\mathcal{N}(\boldsymbol{\mu}_{k,i}^{old}, \boldsymbol{\sigma}_{k,i}^{old})) \text{\quad(Node-level regularization)},\\
    &\mathcal{L}_{\mathrm{prior}} = \sum_{i\in \{l, e,p,s\}}\bigl\|\mathbf z_i\bigr\|_2^{2} \text{\quad(Prior regularization)}.
\end{aligned}
}
\end{equation}

Here, $\mathcal{N}(\boldsymbol{\mu}_{k,i}^{old}, \boldsymbol{\sigma}_{k,i}^{old})$ denotes the original latent distribution before optimization, $\mathcal{N}(\boldsymbol{\mu}', \boldsymbol{\sigma}')$ represents computed target distribution, and $\{i|r_i<\tau_{o}\}$ denotes the alone nodes in $M$ as computed in Eq.~\eqref{eq:sinkhorn}, where $\tau_{o}=0.1$ .

\subsection{Details of Experimental Setups}\label{app:exp_set}

\subsubsection{Baselines}\label{app:baselines}
We compare four material generative models and six LLMs, including two reasoning-focused LLMs. 
\paragraph{Generative Models.} 
\begin{itemize}
    \item CDVAE~\citep{CDVAE}: A variational autoencoder (VAE)-based model for crystal generation, imposing periodic boundary constraints to capture lattice invariance. It encodes both fractional coordinates and lattice vectors, enabling physically valid crystalline outputs.  
    \item DiffCSP~\citep{diffcsp}: A diffusion model (DM) that incorporates $SE(3)$-equivariant constraints over lattices and fractional coordinates, ensuring rotational and translational invariance during crystal structure prediction.  
    \item SyMat~\citep{SyMat}: A VAE-based framework with symmetry-aware constraints that explicitly enforce geometric symmetry in periodic metamaterials, thereby improving validity under symmetry-preserving transformations.  
    \item Cond-CDVAE~\citep{condCDVAE}: A conditional extension of CDVAE that integrates property vectors into the generative process, enabling structure generation conditioned on target physical properties while preserving periodicity.  
\end{itemize}
\paragraph{Large Language Models (LLMs).}
\begin{itemize}
    \item GPT-4o-mini~\citep{gpt-4o}: A lightweight multimodal variant of GPT-4, optimized for efficiency (tens of billions of parameters), capable of text–vision understanding but not a deep-thinking model.  
    \item Llama-4-maverick~\citep{touvron2023llama}: An open-weight LLM with around 70B parameters, trained for general reasoning and generation. It is not specialized as a reasoning model but provides balanced accuracy and efficiency.  
    \item Deepseek-chat~\citep{liu2024deepseek}: A chat-optimized conversational LLM (hundreds of billions of parameters) designed for dialogue and general problem solving; not a dedicated reasoning model.  
    \item Qwen3-235b~\citep{qwen2.5}: A 235B-parameter deep-thinking model with chain-of-thought style reasoning abilities, explicitly optimized for multi-step reasoning and complex scientific problem solving.  
    \item Deepseek-Reasoning~\cite{guo2025deepseek}: A reasoning-specialized variant of Deepseek, trained with reinforcement learning to enhance long-chain reasoning. It belongs to the emerging class of deep-thinking LLMs.  
    \item Gemini-2.0-flash-lite~\citep{gemini}: A highly efficient multimodal LLM from Google’s Gemini family (tens of billions of parameters), designed for fast inference across text, vision, and structured data, not explicitly reasoning-focused.  
\end{itemize}

\subsubsection{Metrics}\label{app:metrics}
To evaluate the performance, we first employ validity from two aspects, \ie, symmetries and periodicity ~\citep{SyMat,condCDVAE}. To evaluate the generation diversity, we conduct coverage recall that measures how many structures in the test dataset are covered by the generated structures~\citep{chen2025metamatbench}. In addition, we introduce the repeat ratio to indicate how many of the same structures are generated by one model. The more detailed computation of these metrics is illustrated as follows.
\paragraph{Symmetry Validity.} Symmetry validity is to evaluate the symmetry of a structure by computing the central symmetry ratio ($\mathcal{V}_{S}$) of a graph in the 3D Cartesian space. Specifically, $\mathcal{V}_{S}$ is defined as:
\begin{equation}
    \mathcal{V}_{S} = \frac{1}{N_L}\sum_{k}^{N_L}\frac{N_{S_k}\cdot\sum_i^{N_k}s_{degree_i}}{{N_k}^2},
\end{equation}
where $N_L$ is the number of generated structures, $N_k$ is the node number of $k$-th structure, and $N_{S_k}$ is the number of Symmetrical Node that is defined in Definition.~\ref{def:symmetric_node} in $k$-th structure, and $s_{degree_i}$ denotes Symmetry Degree that is defined in Definition~\ref{def:symm_degree}.
In detail, we define a symmetrical node as a node that can find central symmetrical ones within an error range: 
\begin{definition}[Symmetrical Node]
\label{def:symmetric_node}
        $\mathbf{p}_c$ denotes central coordinates in this structure, and $\epsilon$ is a positive hyperparameter. We consider node $i$ with coordinates $\mathbf{p}_i$ to be a symmetrical node iff exists another node $j$ in the structure satisfies: 
        $
            \left\Vert \mathbf{p}_i + \mathbf{p}_j - 2\mathbf{p}_c\right\Vert_2 < \epsilon.
        $
\end{definition}
\noindent In addition, the symmetry degree of a node is defined as the error value of the corresponding "most symmetric" node pair divided by the distance between the central coordinates and the farthest node.
\begin{definition}[Symmetry Degree]
    \label{def:symm_degree}
        $\mathbf{p}_c$ denotes central coordinates in this structure, and $j$ is a node in this structure. The symmetry degree of node $i$ in a structure is defined as:
           $ s_{degree_i} = \frac{\epsilon_{max} - s_{error_i}}{\epsilon_{max}},$
        where $\epsilon_{max} = \max_j{\Vert \mathbf{p}_c - \mathbf{p}_j \Vert_2}$, and $s_{error_i} = \min_j{ \Vert \mathbf{p}_i + \mathbf{p}_j - 2\mathbf{p}_c\Vert_2}$.
\end{definition}
\paragraph{Periodicity Validity.} According to Definition~\ref{def:lattice}, a lattice is formed by periodically repeating unit cell structures along the lattice vectors $\mathbf{L}$. Therefore, the periodicity, denoted as $\mathcal{V}_P$, aims to assess the generated structures at the lattice level. This metric aims to evaluate whether the structures can repeat for constructing a lattice, 
Formally, we define the necessary condition of periodicity of a structure:
    \begin{definition}[Periodicity]
    \label{def:lattice}
        Given a structure with node positions $\mathbf{P}$ and lattice vectors $\mathbf{L}$, for each dimension $d \in \{0, 1, 2\}$, there exist at least one pair of coordinate points $\mathbf{p}_i$ and $\mathbf{p}_j$ such that $\mathbf{p}_i + \mathbf{l}_d$ is approximately equal to $\mathbf{p}_j$ in the L1 norm within a tolerance range $\epsilon$. Formally,
        \begin{equation*}
            \begin{aligned}
            &\forall d \in \{0, 1, 2\},\\
            &\exists i \in \{0, 1, \ldots, N-1\}, \exists j \in \{0, 1, \ldots, N-1\},\\
            &\text{s.t. } \| (\mathbf{c}_i + \mathbf{l}_d) - \mathbf{c}_j \|_1 < \epsilon.
            \end{aligned}
        \end{equation*}
    \end{definition}
\noindent Eventually, the evaluation of the periodicity of generated lattices can be computed by $\mathcal{V}_P = \frac{N_P}{N_L}$, where $N_P$ denotes the number of generated structures that satisfy Definition~\ref{def:lattice}.

\paragraph{Coverage Recall (Cov. R).} Intuitively, coverage recall measures how many structures in the ground truth dataset are covered by generated structures,\ie,
    \begin{equation}
    \begin{aligned}
    \text{COV}_R=\frac{1}{N_t}\vert \{&i \in [1,\ldots,N_t]: \exists k \in [1,\ldots,N_L],\\
    &D(\mathbf{P}^*_i, \mathbf{P}_k) < \epsilon_{cov} \} \vert,
    \end{aligned}
    \end{equation}
where $D(\mathbf{P}^*_i, \mathbf{P}_k)$ is a distance metrics to evaluate the distance between $i$th structure and $j$th structure. 
\paragraph{Repeat Ratio.} In detail, for all generated structures, we compute their distance $D(\mathbf{P}^*_i, \mathbf{P}_k)$ and regard the $D(\mathbf{P}^*_i, \mathbf{P}_k) < \epsilon$ as matched structures. To avoid repeat count, we constrain that each structure matches once.

In addition, for the validity and diversity of LLMs that require a prompt input, we use the same prompt of LLMs and ours as ``\textit{Design a valid and diverse structure, ensure it satisfies symmetry, and periodicity}''.
\subsubsection{Dataset}\label{app:dataset}
\begin{figure}[h]
    \centering
    \includegraphics[width=\linewidth]{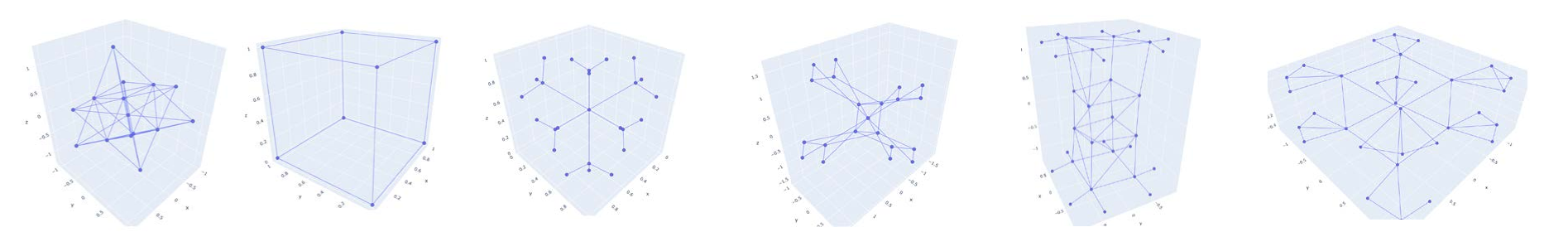}
    \caption{Samples in MetaModulus dataset.}
    \label{fig:data_desc}
\end{figure}

\paragraph{MetaModulus Dataset.}
In this section, we describe the details of MetaModulus~\citep{Modulus} dataset, which contains three mechanical properties, \ie, Young's modulus, Shear modulus, and Poisson's ratio. This dataset provides a comprehensive mechanical illustration for a metamaterial. Specifically, this dataset contains 16,707 samples originally. To filter out some invalid structures with clustered nodes and non-edge nodes, we conduct two criteria in data preprocessing by following previous works~\citep{uniTruss,Modulus,chen2025metamatbench}:(1) filtering out the structure that contains nodes with less than two edges (dangling nodes). (2) filtering out the structure whose node number is more than 100, since generally a metamaterial structure with too many nodes would be unstable and hard to construct in the real world. Finally, the dataset contains a total of 9871 structures, and we select 8000 for training the VAE (Generator) and others for testing the coverage rate (Cov. R). Figure~\ref{fig:data_desc} shows six example structures in this dataset.
\paragraph{Prompts for Language Guidance.}
In order to evaluate the language-guidance effectiveness, we introduce 100 design prompts to test if the model can generate effective structures that fit the prompt semantically. These prompt targets on high-level design concepts, containing terms such as, ``high-stiffness'', ``hard material'', ``extremely flexible'', \etc. Table~\ref{tab:design_prompt} illustrates 10 example prompts from the prompt dataset. We publish the full design prompts data in our code base.
\begin{table}[]
    \centering
    \begin{tabular}{l}
    \toprule
    Design Prompts\\
    \midrule
        Design a structure with high stiffness.\\
        Create a metamaterial with negative Poisson's ratio (auxetic).\\
        Generate a structure with ultra-lightweight and moderate stiffness.\\
        Optimize a design for maximum load-bearing capacity.\\
        Design a structure that is extremely flexible in one direction but rigid in the orthogonal direction.\\
        Build a material with a specific Young's modulus value (e.g., 100 MPa).\\
        Maximize shear strength while minimizing density.\\
        Create a structure with directionally dependent compressive strength.\\
        Design a metamaterial that behaves like a spring under compression.\\
        Create a structure that can absorb large impacts without permanent deformation.\\
        \bottomrule
    \end{tabular}
    \caption{10 samples from all 100 design prompts.}
    \label{tab:design_prompt}
\end{table}
\subsubsection{Implementation Details}
\label{app:training_details}
\begin{figure}
    \centering
    \includegraphics[width=1\linewidth]{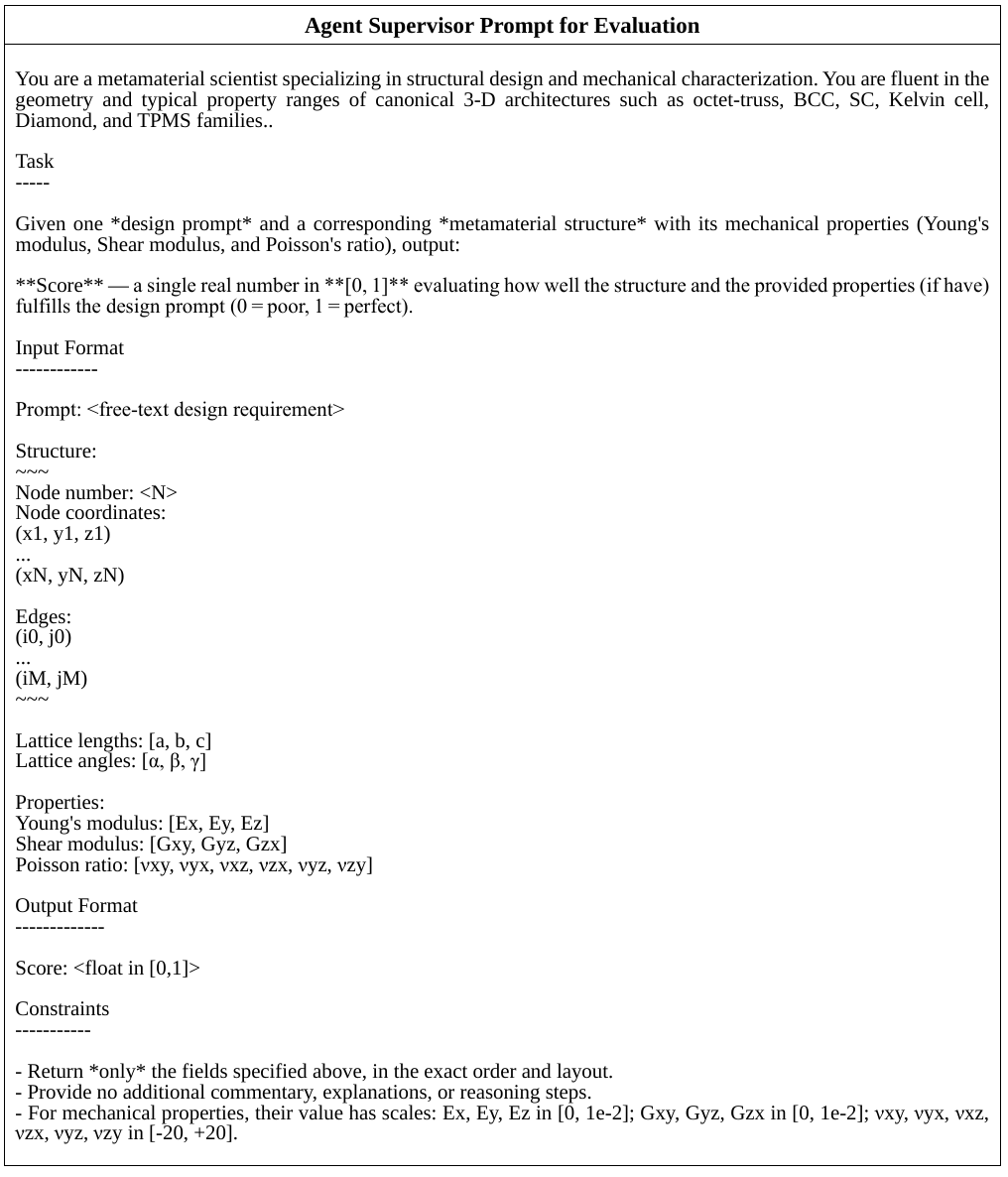}
    \caption{The prompt in Agent Supervisor for evaluation.}
    \label{fig:supervisor_eval}
\end{figure}

\fname~contains three agents, among which there are two architectures required to be trained, \ie, the latent generative model in Agent Generator and the property predictor in Agent Supervisor. All training processes are conducted in NVIDIA A100 or H200 GPUs.
\paragraph{Training Details of Latent Generative Model.} First, we train the VAE according to the disentangled Evidence Lower Bound (ELBO) loss Eq.~\eqref{eq:elbo}. After that, if the instantiation is diffusion model (as depicted in Figure~\ref{fig:enc_dec}), we freeze the VAE and train the denoising model according to the score function Eq.~\eqref{eq:score}. In our experiments, we use VAE as the instantiation rather than the diffusion model. In addition, in the training process, we train the VAE for at most 5000 epochs with early stopping trick on the training data without using validation dataset. We select the checkpoint with the lowest training loss.
\paragraph{Training Details of Predictor in Agent Supervisor.} After the VAE is trained, we incorporate a predictor head of MLPs in the VAE. Specifically, the encoded latents are fed into the predictor head and output the demanded mechanical properties. In the training process, we freeze parameters in the VAE, and only update the parameters in the Predictor head using Mean Squared Error (MSE) loss between prediction and ground truth. We train it for at most 2000 epochs with early stopping strategy. We randomly select 500 samples from training dataset as validation set, and finally use the checkpoint with lowsest MSE in validation set. Moreover, the predictions and ground truth are max-min normalized for a more stable training.
\paragraph{Training Details of Predictor in Agent Supervisor for Evaluation.} In the evaluation process of Section~\ref{sec:quantitative}, we use Supervisor with GPT-4.1 and predictor for scoring. In this phase, the predictor is trained using full dataset, with 500 randomly selected as validation set.
\paragraph{Prompt Details of the LLM in Agent Supervisor for Evaluation.} In the evaluation process of Section~\ref{sec:quantitative}, we use Supervisor with GPT-4.1 as evluator. The overall prompt idea is similar to Figure~\ref{fig:prompt_supervisor}. Differently, we only require it to output scores, without improved properties. Specifically, the prompt in evaluation process is shown in Figure~\ref{fig:supervisor_eval}.

\begin{table*}[h]
\centering
\caption{Convergence analysis. Loss change of iterations.}
\label{tab:convergency}
\begin{tabular}{lccccc}
\toprule
Loss & 0 & 50 & 100 & 150 & 300 \\
\midrule
$\mathcal{L}_s$
  & 28.6
  & 1.85 ($-26.8$)
  & 1.21 ($-0.64$)
  & 0.93 ($-0.28$)
  & 0.78 ($-0.15$) \\
$\mathcal{L}_{p,e}$
  & 147.4
  & 82.2 ($-65.2$)
  & 78.9 ($-3.3$)
  & 81.2 ($+2.3$)
  & 81.9 ($+0.7$) \\
$\mathcal{L}_r$
  & 0.0
  & 123.4 ($+123.4$)
  & 95.8 ($-27.6$)
  & 85.3 ($-10.5$)
  & 76.8 ($-8.5$) \\
$\mathcal{L}_{prior}$
  & 2.22
  & 2.20 ($-0.02$)
  & 2.22 ($+0.02$)
  & 2.24 ($+0.02$)
  & 2.27 ($+0.03$) \\
\bottomrule
\end{tabular}
\end{table*}

\section{More Results}\label{app:results}

\subsection{Convergence Analysis.} \label{sec:convergency}
Table~\ref{tab:convergency} presents the average loss curve of two randomly selected example with the Mix operator, showing that the $\mathcal{L}_s$ and $\mathcal{L}_{p,e}$ decrease until they converge while other two regularization terms $\mathcal{L}_r$ and $\mathcal{L}_{prior}$ tend to converge to a specific value.

\subsection{Validity Experiments for Agent Designer}\label{app:vali_agent1}
To further verify Agent\,1’s ability to produce reasonable scaffolds, we provide additional examples that can be independently checked.

\begin{figure}[t]
    \centering
    \begin{subfigure}{0.45\textwidth}
        \centering
        \includegraphics[width=0.5\linewidth]{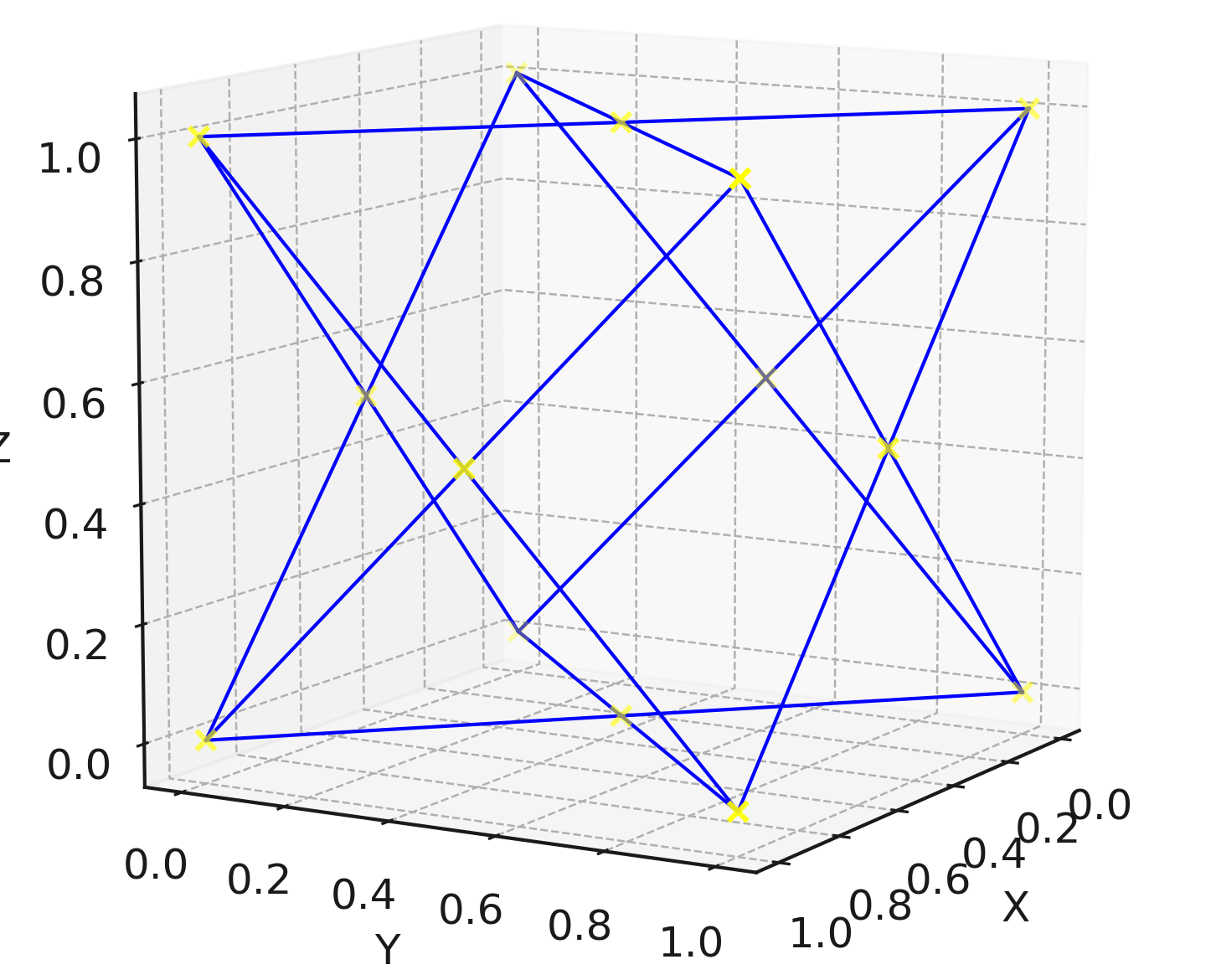}
        \caption{Generated from prompt: Provide a very stable structure with low stiffness.}
    \end{subfigure}
    \hfill
    \begin{subfigure}{0.45\textwidth}
        \centering
        \includegraphics[width=0.5\linewidth]{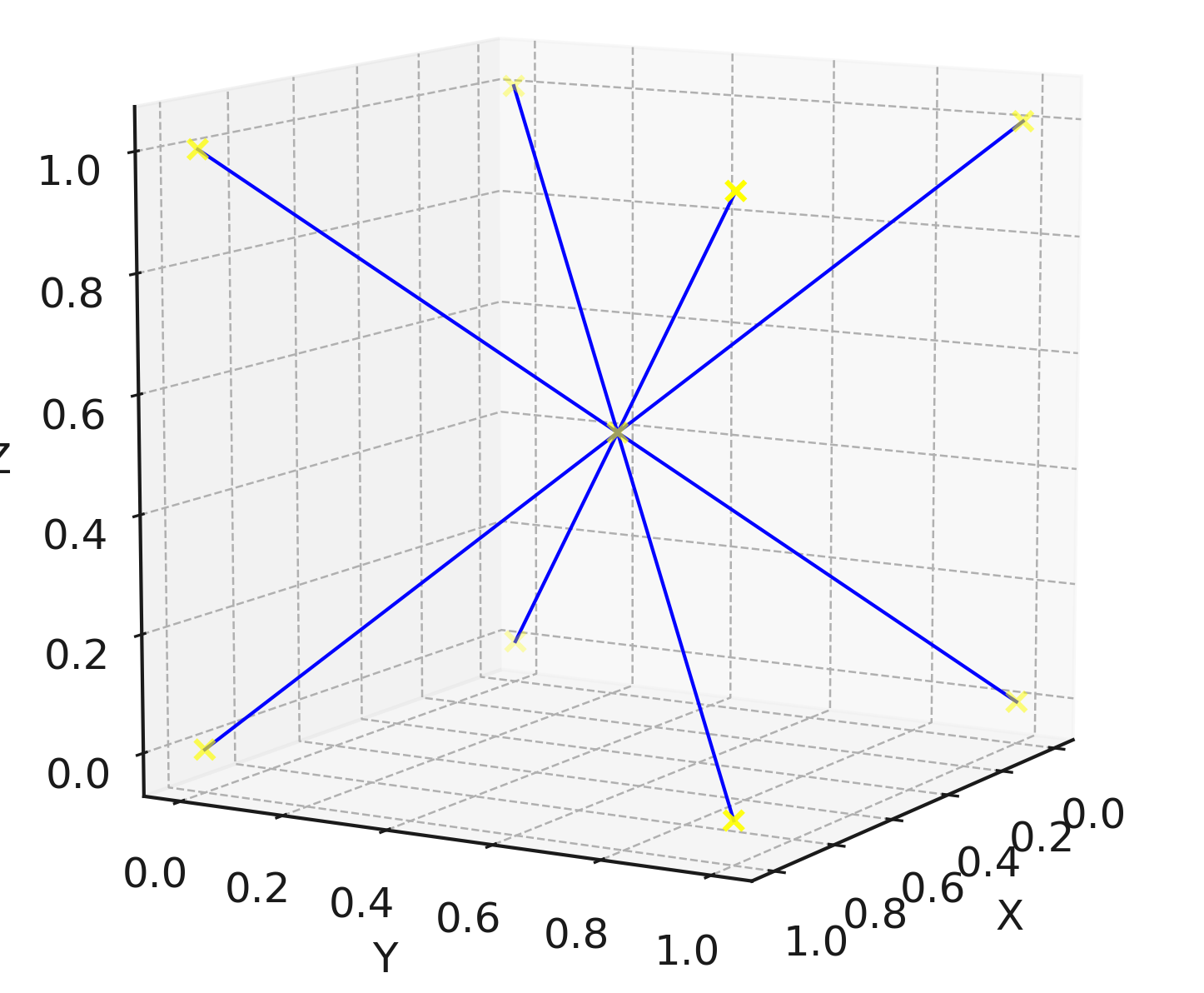}
        \caption{Generated from prompt: Provide a BCC structure.}
    \end{subfigure}
    \caption{Visualization of two generated unit cells produced by Agent Designer.}
    \label{fig:scaffold_examples}
\end{figure}

\begin{itemize}
    \item \textbf{Prompt}: Provide a very stable structure with low stiffness.
\begin{verbatim}
Node number: 14
Node coordinates (fractional):
(0.0, 0.0, 0.0)
(1.0, 0.0, 0.0)
(0.0, 1.0, 0.0)
(1.0, 1.0, 0.0)
(0.0, 0.0, 1.0)
(1.0, 0.0, 1.0)
(0.0, 1.0, 1.0)
(1.0, 1.0, 1.0)
(0.5, 0.5, 0.0)   # face center (z=0)
(0.5, 0.5, 1.0)   # face center (z=1)
(0.5, 0.0, 0.5)  # face center (y=0)
(0.5, 1.0, 0.5)  # face center (y=1)
(0.0, 0.5, 0.5)  # face center (x=0)
(1.0, 0.5, 0.5)  # face center (x=1)

Edges:
(8, 0)
(8, 1)
(8, 2)
(8, 3)
(9, 4)
(9, 5)
(9, 6)
(9, 7)
(10, 0)
(10, 1)
(10, 4)
(10, 5)
(11, 2)
(11, 3)
(11, 6)
(11, 7)
(12, 0)
(12, 2)
(12, 4)
(12, 6)
(13, 1)
(13, 3)
(13, 5)
(13, 7)
\end{verbatim}
    Structural rationale: \\
    High Stability: Triangle crossed struts increase stability. Octet structure is stable. \\
    Low Stiffness: Minimal edge connectivity along principal directions reduces stiffness, providing flexibility and deformation capacity under loading.

    \item \textbf{Prompt}: Provide a BCC structure.
\begin{verbatim}
Node number: 9
coordinates:
(0,0,0)
(1,0,0)
(1,1,0)
(0,1,0)
(0,0,1)
(1,0,1)
(1,1,1)
(0,1,1)
(0.5,0.5,0.5)
Edges:
(0,8)
(1,8)
(2,8)
(3,8)
(4,8)
(5,8)
(6,8)
(7,8)
\end{verbatim}

\end{itemize}
According to the results, we can visualize the scaffold examples as in Figures~\ref{fig:scaffold_examples}.

\subsection{Additional Qualitative Analysis of Intersection and Negation.}\label{app:qual}
\begin{figure}
    \centering
    \includegraphics[width=\linewidth]{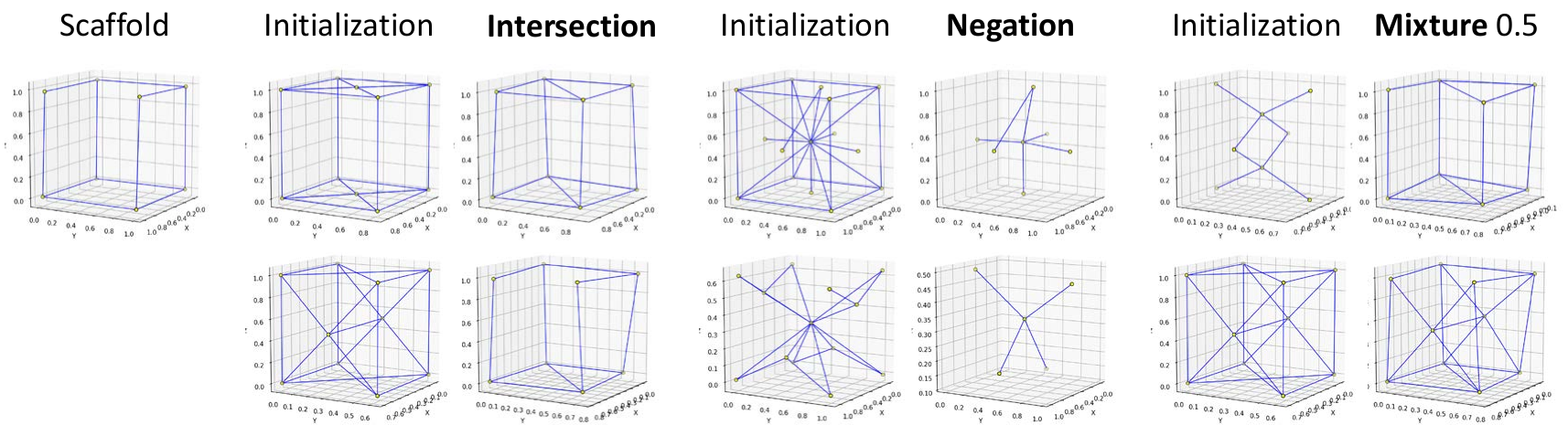}
    \caption{Qualitative results of Intersection, Negation, and Mixture symbolic operators.}
    \label{fig:mix_neg_int}
\end{figure}
We provide additional qualitative results for the Intersection and Negation operators in Figure~\ref{fig:mix_neg_int}. Unlike Union and Mix, these two operators are not intended to produce direct geometric merging or interpolation. Instead, they operate at the semantic level in latent space. Intersection emphasizes the common semantic components shared by the initialization and the scaffold, encouraging the generated structure to retain only their mutually consistent attributes. Negation, by contrast, suppresses scaffold-related latent factors, steering the generation away from the emphasized semantics of the scaffold. As a result, their outputs should not be interpreted as literal geometric intersection or subtraction in Euclidean space, but rather as semantic transformations induced in the disentangled latent space. The qualitative results are consistent with this design: both operators produce meaningful structural changes, while exhibiting less direct controllability than Union and Mix, which is why we focus on the latter two in the main text.
\subsection{More Analysis for Property Predictor in Agent Supervisor}\label{app:more_analy_predictor}
\begin{figure}
    \centering
    \includegraphics[width=1\linewidth]{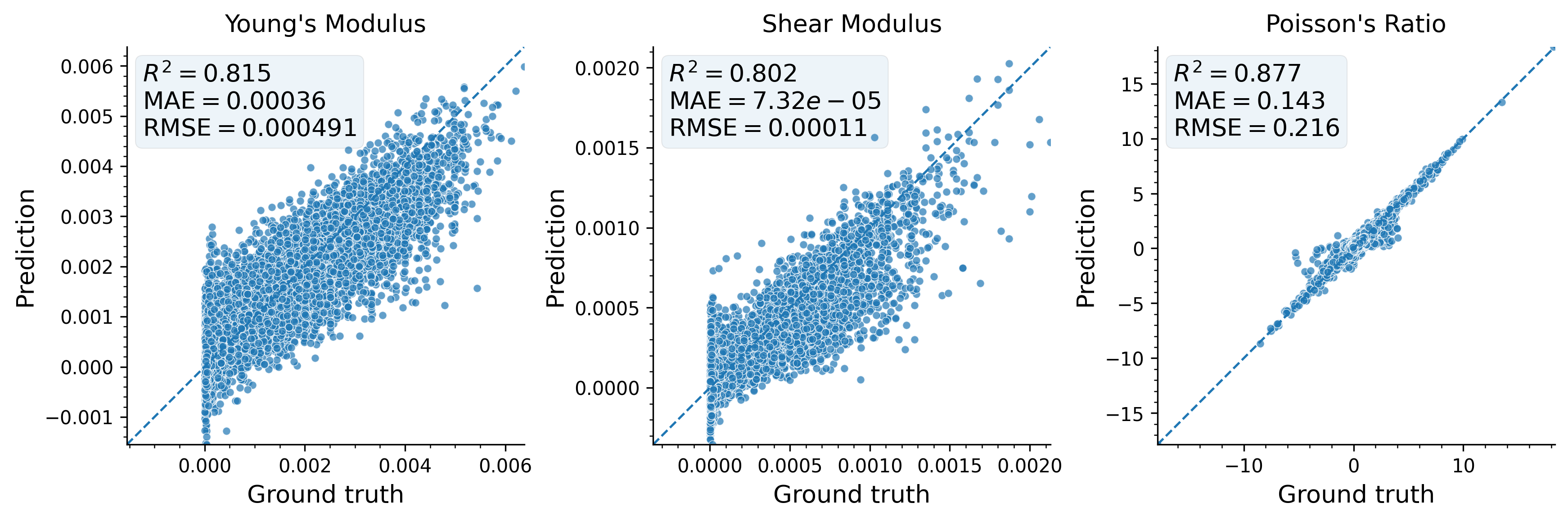}
    \caption{Prediction results of the proposed disentangled VAE on three properties. All $R^2 > 0.8$ demostrates the strong prediction results on the three mechanical properties.}
    \label{fig:all_prop_pred}
\end{figure}

Figure~\ref{fig:all_prop_pred} shows the fitting performance of the proposed disentangled predictor on the test set regarding three mechanical properties (\ie, Young's modulus, Shear modulus, and Poisson's ratio). It can be seen that all fitting $R^2 > 0.8$, demonstrating effective fitting performance. Mover, Table~\ref{tab:mae_comparison} compared our proposals with several existing works, includding invariant model SphereNet~\citep{spherenet}, Equivariant model~\citep{equiformer} and ViSNet~\citep{visnet}, and valina VAE, demonstrating the superiority of disentangled semantic latent in effectively capturing mechanical proeprties.
\begin{table}[tbp]
\small
\centering
\caption{Performance comparison of different models on predicting mechanical properties (MAE). Show the superior performance of the proposed framework.}
\begin{tabular}{lccc}
\hline
Models & Young's Modulus & Shear Modulus & Poisson's Ratio \\
\hline
Vanilla VAE & $6.3 e^{-4}$ & $1.4 e^{-4}$ & 0.39 \\
SphereNet~\citep{spherenet} & $4.7 e^{-4}$ & $1.0 e^{-4}$ & 0.35 \\
Equiformer~\citep{equiformer} & $6.6 e^{-4}$ & $2.2 e^{-4}$ & 0.36 \\
ViSNet~\citep{visnet} & $6.2 e^{-4}$ & $6.3 e^{-2}$ & 0.37 \\
MetaSymbO (disentangled VAE) & $\boldsymbol{3.5 e^{-4}}$ & $\boldsymbol{7.3 e^{-5}}$ & \textbf{0.14} \\
\hline
\end{tabular}
\label{tab:mae_comparison}
\end{table}
\subsection{More Case Studeis}\label{app:case_studies}
\begin{figure}[h]
    \centering
    \includegraphics[width=\linewidth]{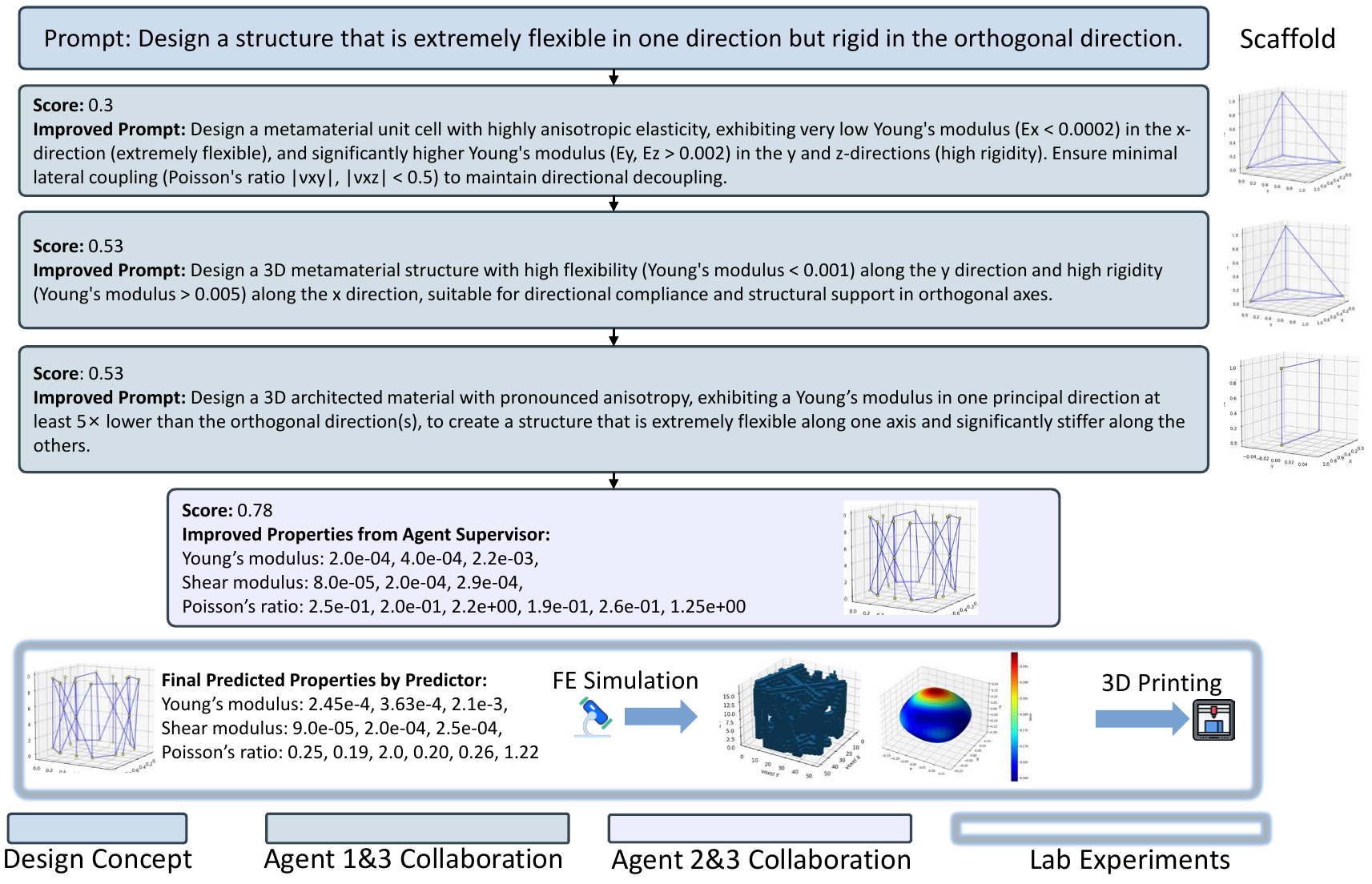}
    \caption{Second case study.}
    \label{fig:case_study2}
\end{figure}

\begin{figure}[h]
    \centering
    \includegraphics[width=\linewidth]{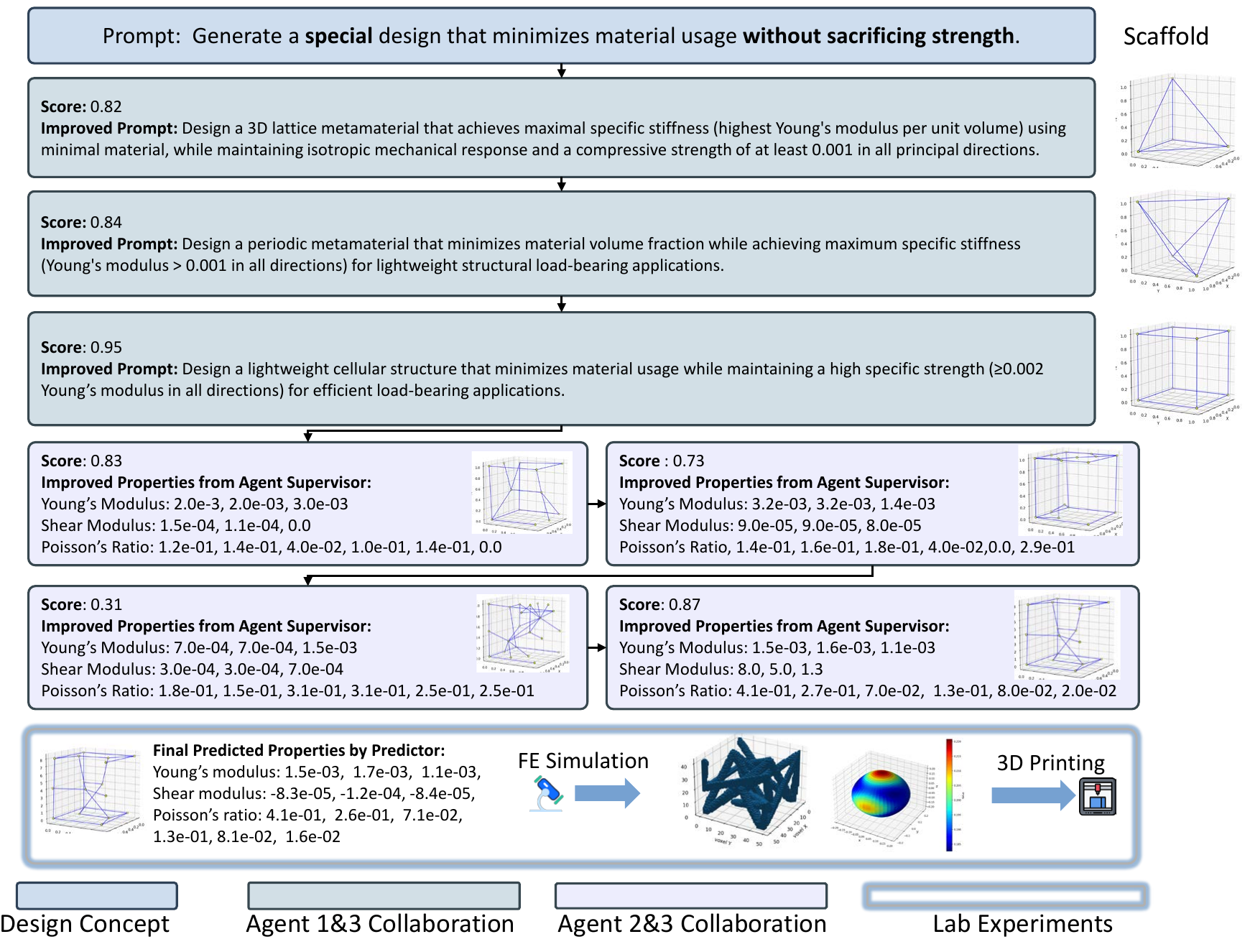}
    \caption{Third case study.}
    \label{fig:case_study3}
\end{figure}

In this section, we show more case studies given a design concept.
Figures~\ref{fig:case_study2} and ~\ref{fig:case_study3} show two other case studies, which demonstrate the effectiveness of multi-agent collaborations. Specifically, Case study 2 (Figure~\ref{fig:case_study2}) proposes a simple scaffold that has extreme flexibility at first, and combines it with a possible structure that contains the most suitable mechanical properties to generate final results. In addition, Case study 3 (Figure~\ref{fig:case_study3}) shows that the evaluation score increases with multi-agent collaboration iterations. Finally, it generates a complicated but suitable structure with high density, which implies high rigidity.

\subsection{Wet Lab Results}
\begin{figure}[t]
    \centering
    \begin{subfigure}[b]{0.28\linewidth}
        \centering
        \includegraphics[width=\linewidth]{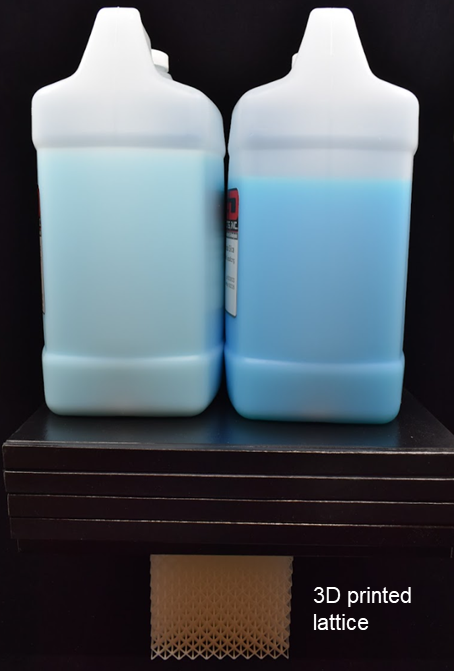}
        \caption{3D printing and simulation example.}
        \label{fig:3d_examples}
    \end{subfigure}
    \hfill
    \begin{subfigure}[b]{0.68\linewidth}
        \centering
        \includegraphics[width=\linewidth]{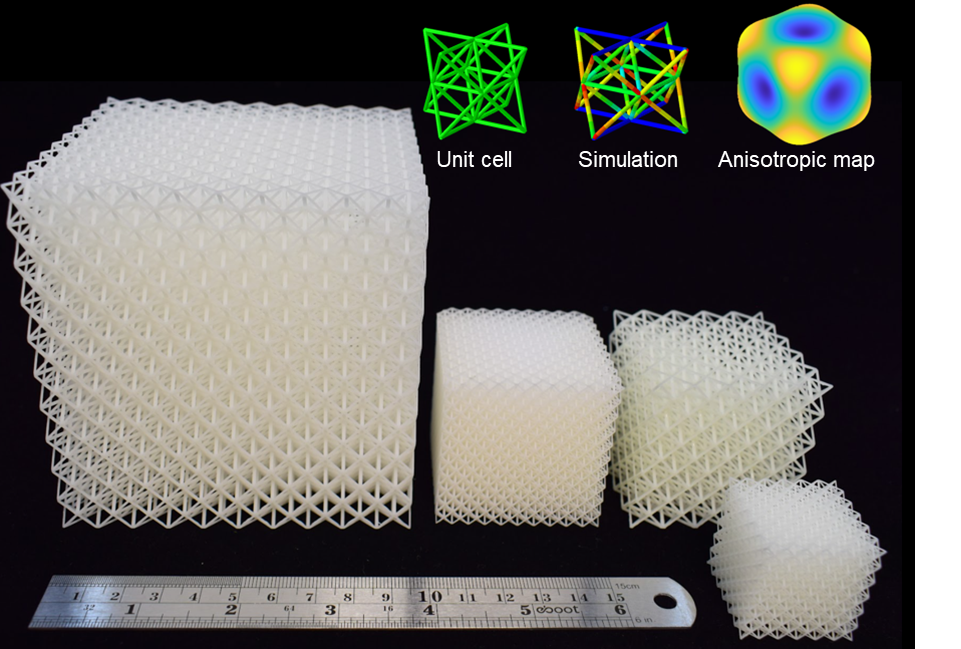}
        \caption{3D-printed lattice samples.}
        \label{fig:3d_lattice}
    \end{subfigure}
    \caption{Wet-lab validation via 3D printing. Left: a proof-of-concept printed lattice with simulation context. Right: representative 3D-printed lattice samples at different scales, demonstrating the manufacturability of the generated periodic structures.}
    \label{fig:3d_printing_results}
\end{figure}

For wet-lab validation, Figures~\ref{fig:3d_examples} and \ref{fig:3d_lattice} present representative 3D-printed lattice specimens obtained by periodically tiling the generated truss-like unit cells into macroscopic samples, consistent with the lattice-based metamaterial formulation of the framework.

Figure~\ref{fig:3d_examples} shows a small proof-of-concept print placed next to common laboratory containers, providing an intuitive reference for the physical scale of the fabricated structure and the fine resolution of its strut network. Figure~\ref{fig:3d_lattice} further shows several larger printed samples with different overall sizes and shapes, together with the corresponding unit-cell, simulation, and anisotropy visualizations, illustrating the full pipeline from digital design to physical realization. The preserved open-cell topology, regular repetition of unit cells, and visually clean node--edge connectivity indicate that the generated structures are practically printable, supporting the use of 3D printing as a wet-lab validation step for the final designs in our case study.

\section{Theoretical Analysis}\label{app:theoretical}
\subsection{Latent Generation Models}\label{app:latent_models}
Auto-encoder (AE) is widely used in the material discovery domain to compress discrete material data to a continuous latent space for downstream tasks (especially inverse material design) \citep{zeng2024data,hanakata2020forward}. VAEs~\citep{vae} introduce a Gaussian prior and derive evidence lower bound optimization (ELBO). More recently, latent diffusion models (LDMs) are extensively explored due to its strong ability in reconstructing high-fidelity data~\citep{rombach2022high,podell2023sdxl,fu2024latent}. 
Both VAEs and LDMs typically impose a multivariate Gaussian prior on the latent space, thereby enabling the application of symbolic logic operators directly within the latent Gaussian manifold. 

Therefore, Agent Generator is instantiated as an AE-based latent generation model in this work, and we implement the basic version of VAEs and LDMs for experiments.

Formally, given a metamaterial $M = (\mathbf{L}, \mathcal{U})$, Both VAEs and LDMs operate by encoding $M$ into a continuous latent variable $\mathbf{z} \in \mathcal{Z} \subseteq \mathbb{R}^d$ via a stochastic encoder: $q_{\boldsymbol{\phi}}(\mathbf{z} \mid M)$, and reconstruct the input through a decoder: $p_{\boldsymbol{\theta}}(M \mid \mathbf{z})$.
Similar for VAEs and LDMs, a standard Gaussian prior is imposed over the latent space: $p(\mathbf{z}) = \mathcal{N}(\mathbf{z} \mid \mathbf{0}, \mathbf{I})$. The difference lies in the objectives. Specifically, VAEs conduct ELBO for optimization:
\begin{equation}
\label{eq:org_elbo}
    \mathcal{L}_{\text{VAE}}(\boldsymbol{\phi}, \boldsymbol{\theta}) = \mathbb{E}_{q_{\boldsymbol{\phi}}(\mathbf{z} \mid M)} \left[ \log p_{\boldsymbol{\theta}}(M \mid \mathbf{z}) \right] 
- \mathrm{KL}\left( q_{\boldsymbol{\phi}}(\mathbf{z} \mid M) \,\|\, p(\mathbf{z}) \right),
\end{equation}
while in LDMs, latent variables $\mathbf{z}_0$ are further corrupted over $T$ steps via a diffusion process and denoising process:
\begin{equation}
    \begin{aligned}
    &q(\mathbf{z}_t \mid \mathbf{z}_0) = \mathcal{N}(\mathbf{z}_t \mid \sqrt{\bar{\alpha}_t} \mathbf{z}_0, (1 - \bar{\alpha}_t) \mathbf{I}),
    \\
    &p_{\boldsymbol{\theta}}(\mathbf{z}_{t-1} \mid \mathbf{z}_t) = \mathcal{N}(\mathbf{z}_{t-1} \mid \boldsymbol{\mu}_{\boldsymbol{\theta}}(\mathbf{z}_t, t), \boldsymbol{\Sigma}_{\boldsymbol{\theta}}(t)),
    \end{aligned}
\end{equation}
where $\bar{\alpha}_t = \prod_{s=1}^{t} \alpha_s$, and $\{\alpha_t\}$ is a predefined noise schedule. The training objective minimizes the expected noise prediction to learn the parameters $\theta$ in denoising model
$
\mathcal{L}_{\text{LDM}}(\boldsymbol{\theta}) = \mathbb{E}_{\mathbf{z}_0, \boldsymbol{\epsilon}, t} 
\left\| \boldsymbol{\epsilon} - \boldsymbol{\epsilon}_{\boldsymbol{\theta}}(\mathbf{z}_t, t) \right\|^2
$,where $\mathbf{z}_t = \sqrt{\bar{\alpha}_t} \mathbf{z}_0 + \sqrt{1 - \bar{\alpha}_t} \boldsymbol{\epsilon}$, and $\boldsymbol{\epsilon} \sim \mathcal{N}(\mathbf{0}, \mathbf{I})$ is standard Gaussian noise.

\subsection{Derivation of Symbolic Operators in Gaussian}\label{app:gaussian_deriv}
\subsubsection{Derivation of Mix Operator (Eq.~\eqref{eq:mixture})}\label{app:mix_deriv}
Given two diagonal-Gaussian latent posteriors
\(
q_M(\mathbf{z}) = \mathcal{N}\!\bigl(\boldsymbol{\mu}_M,\operatorname{diag}\boldsymbol{\sigma}_M^{2}\bigr)
\)
and
\(
q_{M'}(\mathbf{z}) = \mathcal{N}\!\bigl(\boldsymbol{\mu}_{M'},\operatorname{diag}\boldsymbol{\sigma}_{M'}^{2}\bigr),
\)
their convex combination is
\begin{equation}
p_{\mathrm{mix}}(\mathbf{z})
      =(1-\lambda_{mix})\,q_M(\mathbf{z}) + \lambda_{mix}\,q_{M'}(\mathbf{z}),
\qquad
\lambda_{mix}\in[0,1].
\label{eq:raw_mixture}
\end{equation}

\[
\mathbb{E}_{p_{\mathrm{mix}}}[\mathbf{z}]
      =(1-\lambda_{mix})\boldsymbol{\mu}_M + \lambda_{mix}\boldsymbol{\mu}_{M'}
      =\boldsymbol{\mu}_{\mathrm{mix}}.
\]

Let
\(\boldsymbol{\Sigma}_M=\operatorname{diag}\boldsymbol{\sigma}_M^{2}\) and
\(\boldsymbol{\Sigma}_{M'}=\operatorname{diag}\boldsymbol{\sigma}_{M'}^{2}\).
Then
\begin{equation}
\boldsymbol{\Sigma}_{\mathrm{mix}}
   =(1-\lambda_{mix})\boldsymbol{\Sigma}_M
     +\lambda_{mix}\boldsymbol{\Sigma}_{M'}
     +\lambda_{mix}(1-\lambda_{mix})
       (\boldsymbol{\mu}_M-\boldsymbol{\mu}_{M'})
       (\boldsymbol{\mu}_M-\boldsymbol{\mu}_{M'})^{\top}.
\label{eq:true_moment_match}
\end{equation}

To maintain diagonal structure and avoid the expensive cross term, we drop the outer-product term and interpolate standard deviations:
\[
\sigma_{\mathrm{mix},k}
      \approx (1-\lambda_{mix})\,\sigma_{M,k} + \lambda_{mix}\,\sigma_{M',k}.
\]

With \(\boldsymbol{\sigma}_{\mathrm{mix}}\) defined above and a small constant \(\varepsilon\) for numerical stability,
\begin{equation}\label{eq:mixture_surrogate}
p_{\mathrm{mix}}(\mathbf{z}\mid\lambda_{mix})
  =\mathcal{N}\!\Bigl(
     \mathbf{z}\,;\;
     (1-\lambda_{mix})\boldsymbol{\mu}_M + \lambda_{mix}\boldsymbol{\mu}_{M'},
     \operatorname{diag}\!\bigl((\,(1-\lambda_{mix})\boldsymbol{\sigma}_M + \lambda_{mix}\boldsymbol{\sigma}_{M'}\,)^{2} + \varepsilon\bigr)
  \Bigr)
\end{equation}

\paragraph{Remark.}
Equation~\ref{eq:mixture_surrogate} is a \emph{heuristic single-Gaussian approximation}:
it preserves the exact mean but underestimates the true covariance by omitting  
the cross term in~\ref{eq:true_moment_match}.  This trade-off yields a numerically stable,
fully differentiable latent representation while retaining controllable guidance via~\(\lambda_{mix}\).

\subsubsection{Derivation of Intersection Operator (Eq.~\eqref{eq:intersection})}\label{app:intersection_deriv}
Recall the formula of PoE (Eq.~\eqref{eq:prod}) and the Gaussian distribution of $p_M$ and $p_{M'}$, we have the following derivation by following~\citep{poe,poe_driviation}.
\begin{align*}
p_{\text{int}}(\mathbf{z}) 
&\propto p_M(\mathbf{z})\,p_{M'}(\mathbf{z})  \\[4pt]
&= 
\underbrace{\frac{1}{(2\pi)^{d/2}|\boldsymbol{\Sigma}_M|^{1/2}}
\exp\!\Bigl[-\tfrac12(\mathbf{z}-\boldsymbol{\mu}_M)^\top\boldsymbol{\Sigma}_M^{-1}(\mathbf{z}-\boldsymbol{\mu}_M)\Bigr]}_{\displaystyle p_M(\mathbf{z})}
\;
\underbrace{\frac{1}{(2\pi)^{d/2}|\boldsymbol{\Sigma}_{M'}|^{1/2}}
\exp\!\Bigl[-\tfrac12(\mathbf{z}-\boldsymbol{\mu}_{M'})^\top\boldsymbol{\Sigma}_{M'}^{-1}(\mathbf{z}-\boldsymbol{\mu}_{M'})\Bigr]}_{\displaystyle p_{M'}(\mathbf{z})} \\[8pt]
&\propto 
\exp\!\Bigl[
-\tfrac12\Bigl(\mathbf{z}^\top(\boldsymbol{\Sigma}_M^{-1}+\boldsymbol{\Sigma}_{M'}^{-1})\mathbf{z}
-2\mathbf{z}^\top(\boldsymbol{\Sigma}_M^{-1}\boldsymbol{\mu}_M+\boldsymbol{\Sigma}_{M'}^{-1}\boldsymbol{\mu}_{M'})\Bigr)
\Bigr].
\end{align*}

Let the \emph{precision} matrix be 
\(
\boldsymbol{\Lambda}_{\text{int}}
= \boldsymbol{\Sigma}_M^{-1}+\boldsymbol{\Sigma}_{M'}^{-1}
\)
and define
\(
\boldsymbol{\eta}_{\text{int}}
= \boldsymbol{\Sigma}_M^{-1}\boldsymbol{\mu}_M+\boldsymbol{\Sigma}_{M'}^{-1}\boldsymbol{\mu}_{M'}.
\)
Completing the square gives
\[
p_{\text{int}}(\mathbf{z})
\propto
\exp\!\Bigl[-\tfrac12(\mathbf{z}-\boldsymbol{\mu}_{\text{int}})^\top
\boldsymbol{\Lambda}_{\text{int}}
(\mathbf{z}-\boldsymbol{\mu}_{\text{int}})\Bigr],
\qquad
\boldsymbol{\mu}_{\text{int}}=\boldsymbol{\Lambda}_{\text{int}}^{-1}\boldsymbol{\eta}_{\text{int}}.
\]

Since \(\boldsymbol{\Sigma}_{\text{int}}=\boldsymbol{\Lambda}_{\text{int}}^{-1}\),
we arrive at
\[
p_{\mathrm{int}}(\mathbf{z})=\mathcal{N}\bigl(
\mathbf{z}\,;\,
\boldsymbol{\mu}_{\mathrm{int}},
\boldsymbol{\Sigma}_{\mathrm{int}}
\bigr),
\quad
\boldsymbol{\Sigma}_{\mathrm{int}}=(\boldsymbol{\Sigma}_M^{-1}+\boldsymbol{\Sigma}_{M'}^{-1})^{-1},
\quad
\boldsymbol{\mu}_{\mathrm{int}}=\boldsymbol{\Sigma}_{\mathrm{int}}
\bigl(
\boldsymbol{\Sigma}_M^{-1}\boldsymbol{\mu}_M+
\boldsymbol{\Sigma}_{M'}^{-1}\boldsymbol{\mu}_{M'}
\bigr).
\]

\paragraph{Remark.}  The intersection (PoE) weights each mean by its precision, yielding a sharper Gaussian concentrated where both experts agree.

\subsubsection{Derivation of Negation Operator (Eq.~\eqref{eq:neg})}\label{app:negation_deriv}
Similar to the derivation of PoE~\citep{poe,poe_driviation}, we can derive the following for the negation operator.

Remember that the original and scaffold latent distributions are two multivariate Gaussians
\[
p_{M}(\mathbf{z})=\mathcal{N}\!\bigl(\boldsymbol{\mu}_{M},\boldsymbol{\Sigma}_{M}\bigr), 
\quad 
p_{M'}(\mathbf{z})=\mathcal{N}\!\bigl(\boldsymbol{\mu}_{M'},\boldsymbol{\Sigma}_{M'}\bigr).
\]
We suppress the density peaks of \(M'\) inside \(M\) by defining
\[
p_{\text{neg}}(\mathbf{z})
\;\propto\;
\frac{p_{M}(\mathbf{z})^{\alpha}}
     {p_{M'}(\mathbf{z})^{\beta}},
\qquad
\alpha,\beta>0.
\]
 
Because the log-density of any Gaussian is quadratic, we write
\[
\begin{aligned}
\log p_{\text{neg}}(\mathbf{z})
&= 
\alpha\log p_{M}(\mathbf{z})
-\beta\log p_{M'}(\mathbf{z})
+\text{const}\\[2pt]
&= 
-\tfrac12\mathbf{z}^{\!\top}
\!\bigl(\alpha\boldsymbol{\Sigma}_{M}^{-1}-\beta\boldsymbol{\Sigma}_{M'}^{-1}\bigr)
\mathbf{z}
+\mathbf{z}^{\!\top}
\bigl(\alpha\boldsymbol{\Sigma}_{M}^{-1}\boldsymbol{\mu}_{M}
      -\beta\boldsymbol{\Sigma}_{M'}^{-1}\boldsymbol{\mu}_{M'}\bigr)
+\text{const}.
\end{aligned}
\]  
After that, to ensure the Gaussian, we conduct the moment-matched Gaussian approximation. Specifically, treat the quadratic form above as the (unnormalised) log of a new Gaussian.
Define the \emph{negation precision}
\[
\boldsymbol{\Lambda}_{\text{neg}}
\,\triangleq\,
\alpha\boldsymbol{\Sigma}_{M}^{-1}-\beta\boldsymbol{\Sigma}_{M'}^{-1},
\quad
(\,\boldsymbol{\Lambda}_{\text{neg}}\succ0\ \text{required}\!),
\]
and invert it to obtain the covariance
\(
\boldsymbol{\Sigma}_{\text{neg}}=\boldsymbol{\Lambda}_{\text{neg}}^{-1}.
\)

Multiplying the linear term by \(\boldsymbol{\Sigma}_{\text{neg}}\) gives the mean:
\[
\boldsymbol{\mu}_{\text{neg}}
=\boldsymbol{\Sigma}_{\text{neg}}
\bigl(\alpha\boldsymbol{\Sigma}_{M}^{-1}\boldsymbol{\mu}_{M}
      -\beta\boldsymbol{\Sigma}_{M'}^{-1}\boldsymbol{\mu}_{M'}\bigr).
\]

Finally, we obtain the approximated Gaussian as
\[
p_{\text{neg}}(\mathbf{z})
\;\approx\;
\mathcal{N}\!\bigl(
\mathbf{z};\,
\boldsymbol{\mu}_{\text{neg}},
\boldsymbol{\Sigma}_{\text{neg}}
\bigr),
\quad
\boldsymbol{\Sigma}_{\text{neg}}^{-1}
=\alpha\boldsymbol{\Sigma}_{M}^{-1}-\beta\boldsymbol{\Sigma}_{M'}^{-1}.
\]

The approximation is valid only when 
\(
\alpha\boldsymbol{\Sigma}_{M}^{-1}\succ\beta\boldsymbol{\Sigma}_{M'}^{-1},
\)
ensuring \(\boldsymbol{\Lambda}_{\text{neg}}\) (and hence \(\boldsymbol{\Sigma}_{\text{neg}}\)) is positive definite.

\bigskip
\paragraph{Remark.}  
Choosing \(\alpha\!>\!\beta\) or scaling \(\boldsymbol{\Sigma}_{M'}\) slightly upward guarantees the precision matrix remains positive definite, making the negation operator a well-defined Gaussian.

\end{document}